\documentclass[conference]{IEEEtran}
\usepackage{times}

\usepackage[numbers]{natbib}
\usepackage{multicol}
\usepackage[pagebackref=true,breaklinks=true,colorlinks]{hyperref}
\usepackage{graphicx}
\usepackage{wrapfig,lipsum,booktabs}
\usepackage[font=footnotesize,labelfont=bf]{caption}
\usepackage{subcaption}
\usepackage{bbm}
\usepackage{float}
\usepackage{color, soul} 
\usepackage{wrapfig,lipsum,booktabs}
\usepackage{amsmath}
\usepackage[margin=1in]{geometry}
\usepackage{array}
\usepackage[utf8]{inputenc}
\usepackage{amsmath, amsthm, amssymb}
\usepackage{booktabs}
\usepackage{dsfont}
\usepackage[dvipsnames]{xcolor}
\usepackage{diagbox}
\usepackage[ruled,vlined]{algorithm2e}
\usepackage[dvipsnames]{xcolor}
\usepackage{multirow}
\begin{document}

\title{Mesh-based Dynamics with Occlusion Reasoning for Cloth Manipulation}

\author{Author Names Omitted for Anonymous Review. Paper-ID [16]}
\author{Zixuan Huang, Xingyu Lin, David Held \\ Carnegie Mellon University, Pittsburgh PA 15213, USA\\
\{zixuanhu, xlin3, dheld\}@andrew.cmu.edu}

\newcommand{\dave}[1]{\textcolor{blue}{[\textbf{DH:} #1]}}
\newcommand{\xingyu}[1]{\textcolor{red}{[\textbf{xingyu:} #1]}}
\newcommand{\zixuan}[1]{\textcolor{OliveGreen}{[\textbf{zixuan:} #1]}}

\maketitle

\IEEEpeerreviewmaketitle

\begin{abstract}
Self-occlusion is challenging for cloth manipulation, as it makes it difficult to estimate the full state of the cloth. Ideally, a robot trying to unfold a crumpled or folded cloth should be able to reason about the cloth's occluded regions.
We leverage recent advances in pose estimation for cloth to build a system that uses explicit occlusion reasoning to unfold a crumpled cloth. Specifically, we first learn a model to reconstruct the mesh of the cloth. However, the model will likely have errors due to the complexities of the cloth configurations and due to ambiguities from occlusions.  Our main insight is that we can further refine the predicted reconstruction by performing test-time finetuning with self-supervised losses. The obtained reconstructed mesh allows us to use a mesh-based dynamics model for planning while reasoning about occlusions. We evaluate our system both on cloth flattening as well as on  cloth canonicalization, in which the objective is to manipulate the cloth into a canonical pose. Our experiments show that our method significantly outperforms prior methods that do not explicitly account for occlusions or perform test-time optimization.
Videos and visualizations can be found on our \href{https://sites.google.com/view/occlusion-reason/home}{project website}. 
\end{abstract}


\section{Introduction}
Manipulation of clothing has wide applications but remains a challenge in robotics. Cloth has nearly infinite degrees of freedom (DoF), making state estimation difficult and resulting in complex dynamics. Furthermore, cloth is often subject to severe self-occlusions, especially when it is crumpled. 

Due to self-occlusions, prior methods typically rely on visible features for manipulation, such as wrinkles~\cite{sun2013heuristic} or corners~\cite{maitin2010cloth}. 
Data-driven methods have been proposed 
that can learn a dynamics model in the pixel space~\cite{fabric_vsf_2020} or in a latent space~\cite{yan2020learning}. Recently, Lin, Wang, \emph{et al.}~\cite{lin2022learning} proposed to learn a  mesh dynamics model on the observed partial point cloud of a crumpled cloth (VCD).  
However, VCD only considers a graph over the visible points and does not reason explicitly about the occluded regions of the cloth. This simplification sometimes prevents the planner from finding optimal cloth unfolding actions.

\begin{figure}
    \centering
    \includegraphics[width=0.48\textwidth]{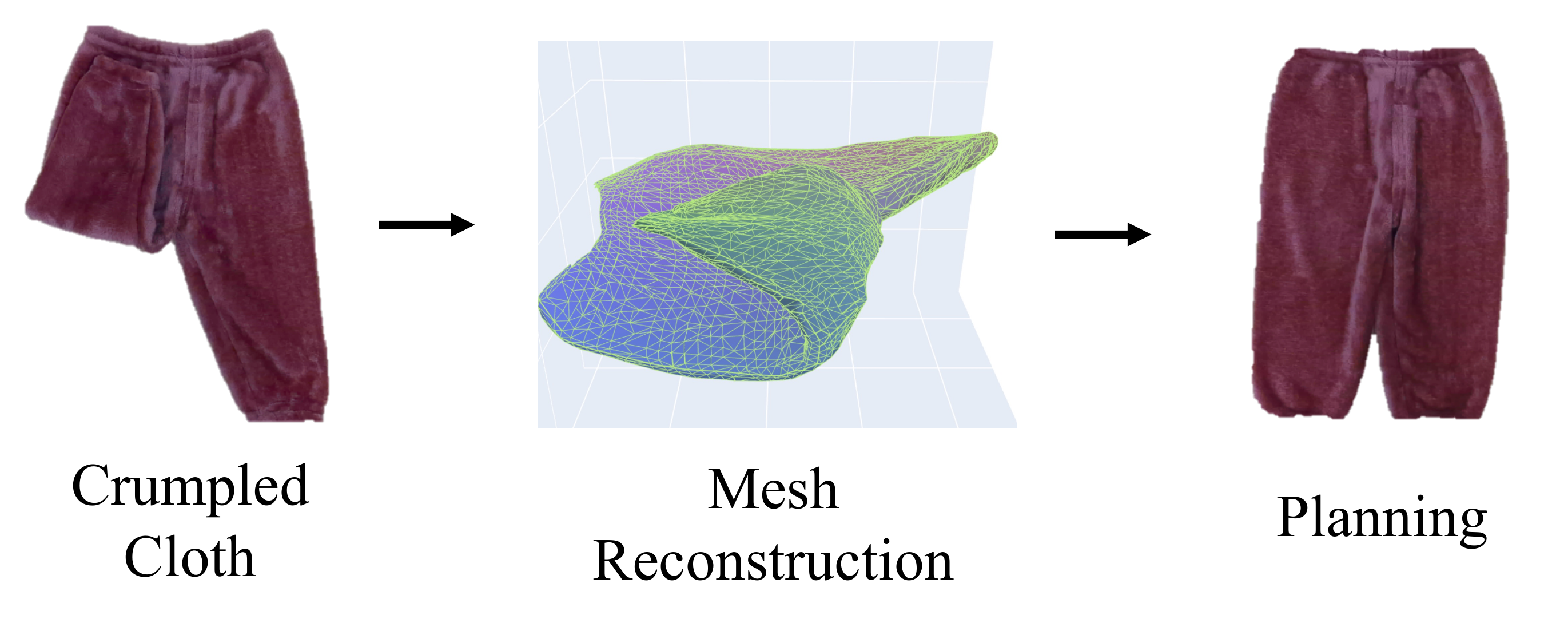}
    \caption{Our method, MEDOR (MEsh-based Dynamics with Occlusion Reasoning), explicitly reasons about occlusions by reconstructing the full mesh of a cloth; we then use a learned mesh-based dynamics model to plan the robot actions.}
    \vspace{-0.5cm}
    \label{fig:teaser}
\end{figure}

Some prior work~\cite{kita2009clothes,li2014recognition,mariolis2015pose,Li2018-sf,li2015folding,li2014real,chi2021garmentnets} attempt to reconstruct the full cloth structure. 
To simplify the task, these approaches typically first grasp and lift the cloth in order to limit the diversity of possible cloth configurations.
However, this approach prevents continual state estimation throughout a manipulation task, since the pose can only be estimated while being held in the air by the gripper at a single point.
In contrast, we aim to estimate the cloth pose from a more diverse set of poses, such as while it is crumpled on a table in arbitrary configurations.
In this paper, we propose MEDOR (MEsh-based Dynamics with Occlusion Reasoning) which is built on top of previous methods for mesh-based dynamics learning~\cite{lin2022learning,pfaff2020learning, sanchez2020learning}.  MEDOR explicitly reasons about occlusions by reconstructing the full cloth mesh from a single depth image.  While previous work such as GarmentNets~\cite{chi2021garmentnets} has demonstrated promising results for mesh reconstruction, we show that, by itself, it is not robust enough for a robot cloth manipulation system. 
 Due to the inherent ambiguity induced by self- occlusions from the crumpled cloth, the reconstructed mesh will likely have many errors, which will create difficulties for downstream planning. Our insight is that we can improve the mesh reconstruction using test-time optimization   with self-supervised losses, which can be easily computed without access to the ground-truth mesh and can be directly optimized on real data. 
 
 We evaluate our method on two tasks: cloth smoothing and cloth canonicalization, which requires aligning the cloth with a canonical flattened pose. We show that our method of mesh reconstruction, with test-time finetuning, enables a robot to smooth or canonicalize the cloth from crumpled configurations more accurately than previous methods. 
 Our contributions include:


\begin{enumerate}
    \item a novel perception model that can better estimate the full cloth structure from crumpled configurations, including a self-supervised test-time optimization procedure. 
    \item a cloth manipulation system that  plans over the reconstructed cloth mesh and can perform both cloth flattening and cloth canonicalization.
\end{enumerate}

\section{Related works}
\subsection{Perception for Cloth Manipulation}
There has been a long history of work on cloth perception for manipulation. We refer to Jimenez \emph{et al.}~\cite{Jimenez2017-xn} for a comprehensive overview. 

Earlier works usually estimate specific visual features of the cloth for manipulation. These include detecting edges and corners for re-grasping~\cite{maitin2010cloth} or grasping and un-folding~\cite{ono1998unfolding,willimon2011model,qian2020cloth}, or detecting wrinkles for smoothing~\cite{sun2013heuristic}. If the cloth is loosely extended, simplified models such as the parameterized shape model~\cite{miller2011parametrized} or the polygonal model~\cite{stria2014polygonal} can also be used for folding. Other works also detect category-specific features such as collars and hemlines~\cite{ramisa2014learning}. These features have been used with hand-designed controllers and strategies; in this work, we aim to learn mesh reconstruction and a cloth dynamics model that we can use to plan a cloth manipulation action sequence.

There are also prior works on more generic pose estimation for clothes. One line of works tries to estimate the pose of on-body clothing from video by leveraging a human body shape prior~\cite{jiang2020bcnet,danvevrek2017deepgarment,patel2020tailornet,saito2021scanimate,pons2017clothcap,hong2021garment4d,su2022deepcloth}. Another line of works assumes the initial configuration of the cloth is known and uses tracking to estimate the full configuration of the cloth under occlusion~\cite{chi2019occlusion, wang2021tracking,tang2018track}. In contrast, we assume that the cloth may be initially crumpled on a table in an unknown initial configuration.

Some other works~\cite{kita2009clothes,li2014recognition,mariolis2015pose,Li2018-sf,li2015folding,li2014real,chi2021garmentnets} simplify the problem by lifting up the cloth using the robot gripper for the purpose of easier pose estimation; lifting up the cloth significantly reduces the set of possible poses and simplifies the reconstruction task. After lifting the cloth, Kita \emph{et al.}~\cite{kita2009clothes} deform a set of predefined representative shapes to fit the observed data. Other works create a dataset of clothes grasped at different locations and retrieve the observed pose at test-time by classification~\cite{li2014recognition,mariolis2015pose,li2015folding,li2014real} or using nearest neighbor~\cite{Li2018-sf}.
In contrast, our method can estimate the cloth directly from a crumpled state on the table, which enables our method to continually re-estimate the cloth state throughout a manipulation sequence.

  
Recently, GarmentNets~\cite{chi2021garmentnets} performed categorical cloth 3D reconstruction by mapping the cloth point cloud into a normalized canonical space (NOCS) defined for each cloth category~\cite{wang2019normalized}. However, like the above methods, GarmentNets also requires grasping and raising the cloth into the air and obtaining four different camera views to reduce the amount of occlusion. In contrast, our perception module only requires a single view of the cloth crumpled on the table.  Further, we find that the reconstructions produced by GarmentNets are not sufficient for accurate cloth manipulation.
Also, unlike GarmentNets that only considers the perception task, we demonstrate a full cloth manipulation system and show the effectiveness of our perception method for manipulation.


\subsection{Data-driven Methods for Cloth manipulation}
Prior works in data-driven cloth manipulation can be categorized as model-free or model-based. Model-based methods train a  policy that outputs actions for cloth manipulation. The policies are trained by either reinforcement learning~\cite{matas2018sim, wu2019learning}, imitation learning~\cite{seita_fabrics_2020}, or by learning a value function~\cite{ha2022flingbot}. Alternatively, the policies can be learned as a  one-step inverse dynamics model~\cite{weng2022fabricflownet,nair2017combining}.

The second approach is to learn a dynamics model and then plan over the model to find the robot actions. Hoque \emph{et al.} directly learn a video prediction dynamics model~\cite{fabric_vsf_2020} and Wilson \emph{et al.} learn a latent dynamics model with contrastive losses~\cite{yan2020learning}. These models do not explicitly reason about the cloth structure, making generalization difficult. Recent work (VCD) trains a mesh dynamics model~\cite{lin2022learning} over the visible points. However, without reasoning about the occluded regions, the planner will often fail to find the optimal smoothing actions. In contrast, our approach plans over a full reconstructed mesh dynamics model.


\subsection{Test-time Optimization}
Test-time optimization has been widely used for view-synthesis~\cite{mildenhall2020nerf,yu2021plenoxels}, 3D particle reconstruction~\cite{chen2021ab} and 3D scene flow~\cite{pontes2020scene}. For example, Chen et al.~\cite{chen2021ab} trained a network to predict an object point cloud that is consistent with a set of object masks in different camera views. 
However, these approaches have not been applied to cloth reconstruction or robot manipulation tasks.

\begin{figure*}[ht]
    \centering
    \includegraphics[width=1\textwidth]{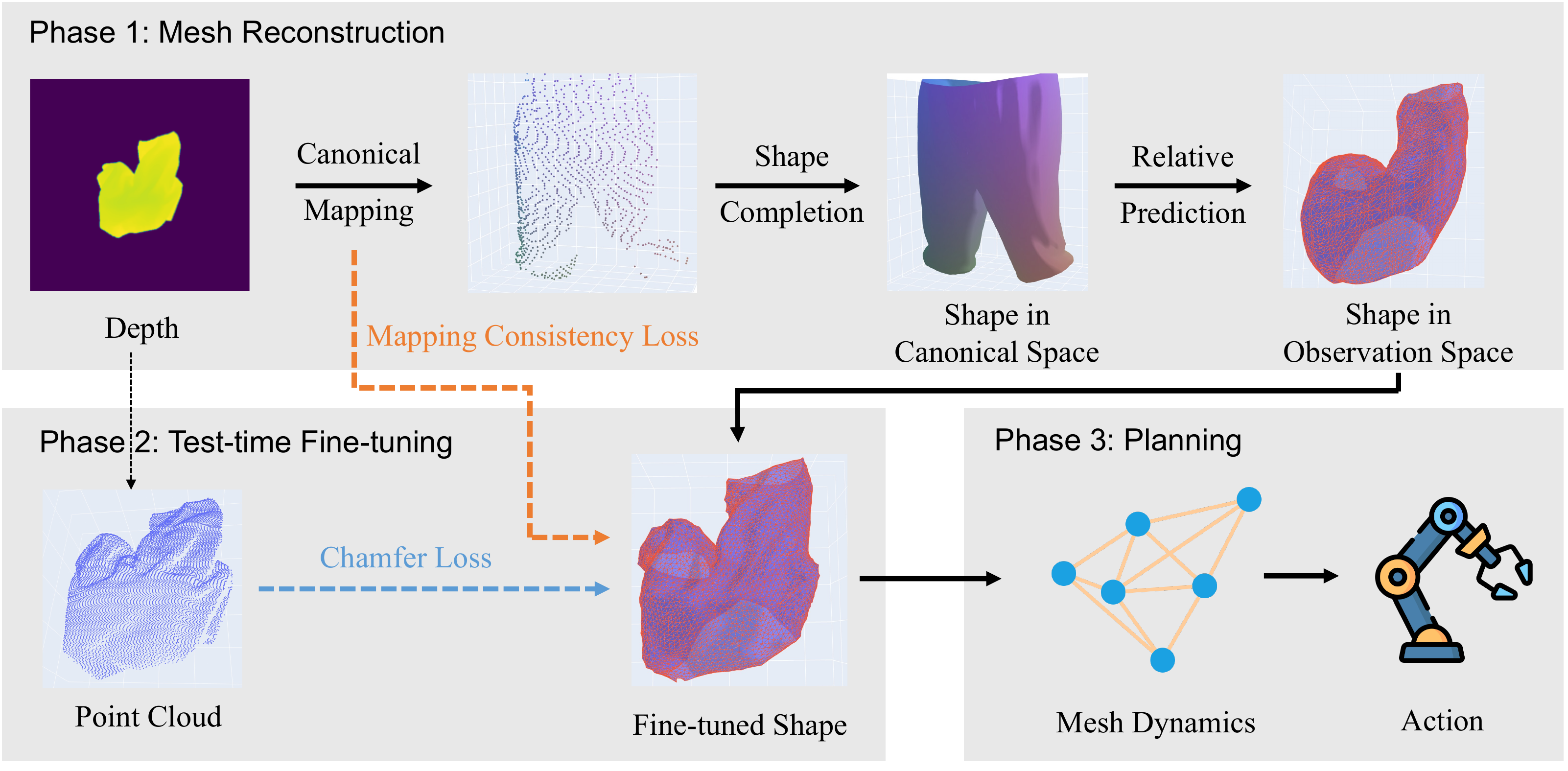}
    \caption{System overview: First, we obtain an initial estimate of the full shape of an observed instance of a cloth from a depth image. In this phase, we obtain an  estimate of the cloth shape in both canonical space and observation space. Then, we conduct test-time finetuning to improve the prediction to better match the observation. Last, we plan with the predicted mesh using a learned mesh-based dynamics model.}
    \label{fig:system_fig}
\end{figure*}

\section{Background}
\subsection{Problem Formulation}
We consider the task of manipulating clothes in a planar workspace with a single robot arm. A cloth at time $t$ is represented by a mesh $M^t=(V^t, E^t)$ with vertices $V^t=\{v_i\}_{i=1...N}$ and mesh edges $E^t$. 
Each vertex consists of a position $x_i$ and velocity $\dot{x_i}$ that will change with the cloth configuration. The ground-truth configuration of the cloth $M^t$ is unknown, and the robot only observes the RGB-D image $I^t$, which includes severe self-occlusions for crumpled garments (though our method only uses the color to segment the cloth from the background). Given the camera intrinsics and the segmentation mask, we can also back-project $I^t$ to a partial point cloud of the cloth. The observation is captured by a top-down camera mounted on the robot end-effector in our setting. As in prior works~\cite{wu2019learning, fabric_vsf_2020, lin2022learning}, we use pick-and-place action primitives for performing cloth manipulation.

\subsection{GarmentNets}
\label{sec:garmentnets}
GarmentNets~\cite{chi2021garmentnets} is a previous work that performs categorical cloth reconstruction from a partial point cloud. 
It contains three steps: first, a normalized canonical space (NOCS~\cite{wang2019normalized}) is defined for each cloth instance by simulating the cloth worn by a human in a T-pose. A canonicalization network (using a PointNet++ architecture~\cite{qi2017pointnet++}) is trained to map an observed partial point cloud to the canonical space (see Figure~\ref{fig:system_fig}, top). 
Second, 
a 3D CNN performs volumetric feature completion to obtain a dense feature grid. An MLP is then used to predict the winding number~\cite{jacobson2013robust} for each point, which is used for surface extraction. In the last step, GarmentNets samples points on the extracted surface and trains a warp field network to map each point in the canonical space back to its location in the observation space.

GarmentNets requires a multi-view depth image to reduce occlusions; it also requires the cloth to be grasped in the air to limit the set of possible poses of the cloth.  In contrast, we consider the more general setup where the cloth is on the table in an arbitrary configuration.

    
    
        
\section{Approach}
Our goal is to build a general cloth manipulation system that can reason about the occluded cloth regions of a cloth from a partial observation and plan robot actions based on these estimates. To do so, we propose a three-part approach: (1) First, given only a partial observation of a crumpled cloth, we train a model to generate a complete mesh of the cloth;
(2) Due to the difficulty of occlusion reasoning, the neural network often fails to accurately reconstruct the cloth mesh. Therefore, we design a test-time finetuning scheme to adapt the predicted mesh to match the observation. (3) Last, we plan with the predicted mesh using a learned dynamics model to find optimal actions for the manipulation task.

\subsection{Estimating the pose of a cloth}
\label{method:coarse_estimate}
Reconstruction of the full cloth structure is fundamentally challenging due to the high-dimensional state space and the ambiguity induced by self-occlusion. As discussed in Section~\ref{sec:garmentnets}, GarmentNets~\cite{chi2021garmentnets} simplifies the problem by using a robot to grasp and lift the cloth and capture 4 observations from different views. In contrast, we tackle the harder problem of estimating the cloth state from a crumpled configuration on the table, so that the cloth state can be re-estimated throughout a manipulation sequence.



To tackle this challenge of cloth pose estimation from crumpled configurations, we make several modifications to GarmentNets that greatly improve its performance:

\noindent\textbf{HRNet:} First, since the cloth is not lifted up, we represent the cloth as a depth image captured by a single top-down camera instead of as a point cloud merged from 4 observations. In this case, the Pointnet++~\cite{qi2017pointnet++} architecture is no longer able to provide reasonable estimates of the cloth configuration, as we will show. Instead, we use an architecture designed for depth images.  In particular, we find that PointNet++ is not able to distinguish whether a piece of cloth is folded above or folded below the rest of the cloth. Differentiating these two cases requires noticing the subtle depth changes at the boundary where two layers meet. Therefore, we replace Pointnet++ with a High Resolution Network (HRNet)~\cite{sun2019deep} which is a convolutional architecture that specializes in producing a high-resolution and spatially precise representation.

\noindent\textbf{Relative predictions:} We found that the reconstruction model sometimes inaccurately estimates the configuration of the cloth in observation space, but it often predicts the canonical shape reasonably well (see Figure~\ref{fig:system_fig} for a visualization of these two spaces). 
Thus, we learn to predict the delta between the position of a point in canonical space and its corresponding location in observation space.
Specifically, given the predicted mesh in canonical space, $\tilde{M}^c=(\tilde{V}^c, \tilde{E}^c), \tilde{V}^c=\{\tilde{v}_i\}_{i=1...n}$, where the coordinate of each vertex in canonical space is $\tilde{x}_i^c$, we predict a 3-dimensional residual vector $\tilde{f}_i$ for each point $i$. The predicted coordinates in the observation space $\tilde{x}_i^o$ are obtained by
\begin{equation}
    \tilde{x}_i^o = \tilde{x}_i^c + \tilde{f}_i
\end{equation}
As shown in the ablation experiments in  Table~\ref{table:ablation}, both modifications described above are important for the performance of the method.


\subsection{Test-time finetuning}
\label{sec:Test-time finetuning}
\begin{figure}
    \includegraphics[width=0.5\textwidth]{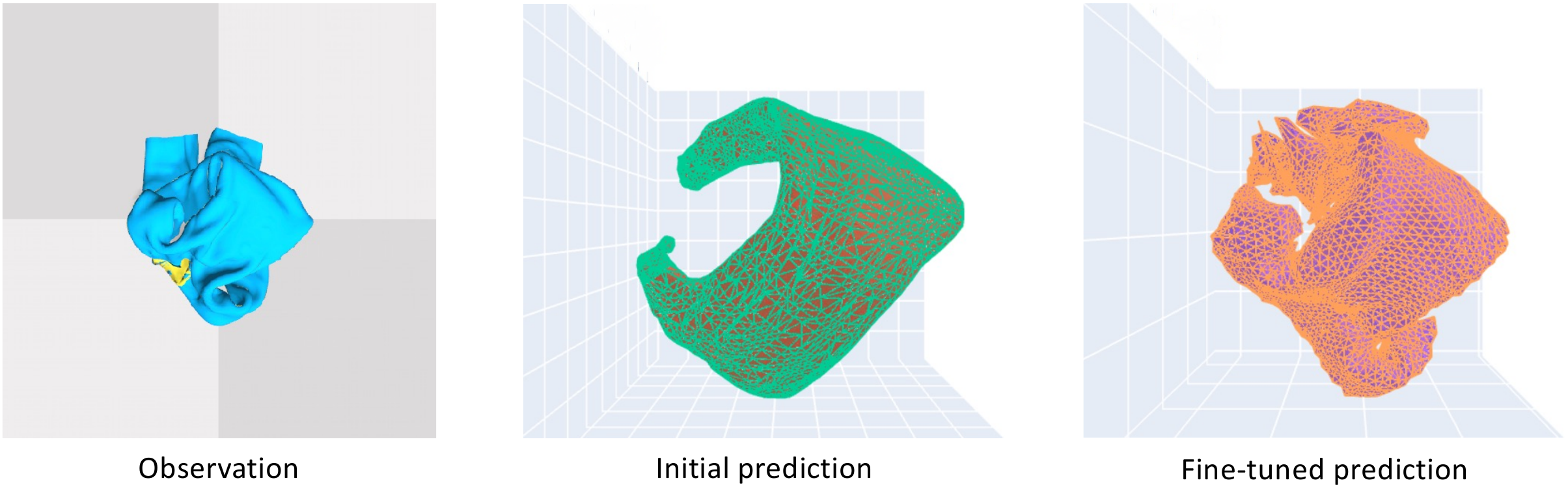}
    \caption{An example showing the benefits of test-time finetuning: The network is able to correctly estimate the main cloth structure, but the initial predicted mesh may be overly smooth (center). Test-time finetuning significantly improves the quality of predicted mesh (right). For more examples, see our \href{https://sites.google.com/view/occlusion-reason/home}{project website}.}
    \vspace{-0.5cm}
    \label{fig:smooth}
    
\end{figure}

Estimating the complete structure of cloth is inherently challenging due to the high degrees of freedom and the ambiguity induced by occlusion. As a result, the network described above still has significant prediction errors, as shown in Fig.~\ref{fig:smooth}.  

To tackle this issue, we design a test-time finetuning scheme that further optimizes the predicted mesh using self-supervised losses that can be computed without knowledge of the ground-truth cloth state. We deform the predicted mesh by optimizing the location of each vertex to optimize an objective consisting of the sum of two loss terms: unidirectional Chamfer loss and mapping consistency loss. 

\noindent\textbf{Unidirectional Chamfer loss}. The first self-supervised loss term 
penalizes any deviations between the predicted mesh and the observed cloth surface~(depth image) so that geometric details are preserved. 
Since the observation only contains information about the visible surface, optimizing the mesh with a standard bi-directional Chamfer loss will result in undesirable results, i.e., the predicted mesh will move entirely to the visible surface and no part of the mesh will remain in the occluded region. Therefore, we use a unidirectional Chamfer loss as described below.

Suppose at timestep $t$, the point cloud observation of the cloth is $P^t=\{p_i\}_{i=1..L}$ and the predicted mesh in observation space is $\tilde{M}^t=(\tilde{V}^t, \tilde{E}^t)$. The coordinate of each vertex is specified by a 3-dimensional vector. 
Then the loss term is formulated as:
\begin{equation}
    \mathcal{L}_\mathcal{C}(\tilde{V}^t; P^t) = \frac{1}{|P^t|}\sum_{p_i\in P^t} \min_{\tilde{v}_j\in \tilde{V}^t} d(p_i, \tilde{v}_j)
\end{equation}
where $d(\cdot, \cdot)$ can be any distance metric, and we use Euclidean distance. In other words, for each point in the observed point cloud, we find the distance to the nearest point in the predicted mesh and minimize the sum of such distances.

\noindent\textbf{Mapping consistency loss}.  Since we don't know the exact correspondence between the predicted mesh and the partial point cloud, only optimizing the uni-directional Chamfer loss above may lead to a local minimum. To alleviate this issue, we observe that the mapping from the observation space to canonical space and then back to observation space (see Figure~\ref{fig:system_fig}, top) creates a cycle; thus we add a loss that, for each visible point, this cycle should end at the location where it started.

Let $P^t$ be the point cloud observation of the cloth; let $f$ be the learned mapping from each observation point to a location in the canonical space; and let $g$ be a learned mapping from each location in the canonical space to a position back in the observation space (shown in Figure~\ref{fig:system_fig}, top).  Note that $g$ operates on the predicted completed cloth surface which includes both observed and occluded points.  The mapping consistency loss term can be expressed as:
\begin{equation}
    \mathcal{L}_\mathcal{M}(P^t) = \frac{1}{|P^t|}\sum_{p_i\in P^t} d(g(f(p_i)), p_i)
\end{equation}
In other words, this loss penalizes the distance between the original location of each observed point $p_i$ and its predicted location $g(f(p_i))$. 

\noindent\textbf{Optimization}.
We use gradient descent with the Adam optimizer~\cite{kingma2014adam} to optimize the losses above. The optimization is divided into two phases. In the first 50 steps, we optimize the mesh using the Chamfer loss together with mapping consistency loss;
then we optimize the mesh using the Chamfer loss alone for another 50 iterations. 

We do not perform a joint optimization throughout the optimization process because multiple pixels in the depth image might be mapped to the same voxel in the canonical space. Enforcing mapping consistency throughout the optimization process will create implausible meshes whose vertices converge to a set of clusters. Instead, we use the mapping consistency loss to provide a good initialization (beyond the initial network prediction), and then we use the uni-directional Chamfer loss to refine the geometric details. As shown in the ablation in Table~\ref{table:ablation}, the two-stage optimization is critical for the performance. 


\subsection{Planning with GNN-based dynamics model}
\label{sec:method:dyn}
Once we have reconstructed the cloth mesh, we use the reconstructed mesh to plan the robot actions using a dynamics model.  One option would be to use a physical simulator as a dynamics model.  However, our experiments show that using a learned dynamics model can be more accurate and faster than a simulator when planning with a reconstructed mesh; these results (shown in Sec.~\ref{sec:exp:ablation}) are consistent with previous papers~\cite{lin2022learning}.  We suspect that a learned dynamics model can be more robust to errors in the mesh reconstruction compared to a physics-based simulator.


As a first step towards learning a dynamics model, we downsample the mesh by vertex clustering~\cite{low1997model} for faster computation. We apply vertex clustering in the canonical space (defined as the pose of the cloth when worn by a human in a T-pose) because the cloth surfaces are well separated and thus will avoid undesirable artifacts such as merging different layers. 

Given a down-sampled mesh from the simulator $\bar{M}^t = (\bar{V}^t, \bar{E}^t)$, we train a dynamics Graph Neural Network (GNN)~\cite{sanchez2020learning, pfaff2020learning}. A GNN encodes the input feature on the nodes and on the edges and then conducts multiple message passing steps between the nodes and edges~\cite{scarselli2008graph}. The decoder will decode the latent features of each node into the predicted acceleration for that node. The GNN dynamics model that we use is the same as the dynamics model in VCD~\cite{lin2022learning}: the input feature on each node is the historic particle velocities 
and an indicator of whether the node is picked by the gripper or not. The edge features includes the distance vector of connected vertices $(x_j - x_k$), its norm $||x_j - x_k||$, and the current displacement from the rest position $||x_j - x_k|| - r_{jk}$. We use Euler integration to obtain the states of the cloth in the next time step. The action is encoded to the dynamics model by directly modifying the position and velocity of the grasped point on the cloth. We refer the reader to previous work~\cite{lin2022learning} for details on the graph dynamics model. Once we have the dynamics model, we use random shooting to plan over different pick-and-place actions and pick the action with the highest predicted reward. At each time step, we plan over a horizon of one, i.e., one pick-and-place action. We train a single dynamics model on Trousers and then apply it to multiple categories in our experiments.

\subsection{Implementation details}
All models are trained in simulation and data are generated by Nvidia Flex wrapped in Softgym~\cite{corl2020softgym}. To obtain a diverse dataset with garments of different sizes and shapes, we port the CLOTH3D dataset~\cite{bertiche2020cloth3d} into Softgym. We choose five categories of garments from CLOTH3D: Trousers, T-shirts, Dress, Skirt and Jumpsuit. Each category contains 400-2000 different meshes and covers a wide range of variations, such as shirts with short and long sleeves, with and without an opening in front. 
We divide the CLOTH3D dataset into a train set and test set in a 9:1 ratio.

\paragraph{Mesh reconstruction model} Initial crumpled cloth configurations are generated by a random drop or random pick-and-place actions from flattened states, with a ratio of 1:1.
The mesh reconstruction model is trained in a category-dependent manner, i.e. one mesh reconstruction model per category, similar to GarmentNets~\cite{chi2021garmentnets}. For each category, the training set contains 20,000 observations.

\paragraph{Dynamics model} The GNN dynamics model is trained only on Trousers and is evaluated across all object categories. The dynamics model is trained on a dataset with 5,000 pick-and-place actions, which equals 500,000 intermediate timesteps. The training of the mesh reconstruction model and the dynamics model each take around 3 days.

At test-time, the mesh reconstruction model and test-time finetuning take around 2.5 and 3 seconds per mesh respectively. For planning, we randomly sample 500 pick-and-place actions and rollout each action with the GNN dynamics model, which takes around 100 seconds. For additional implementation details, please refer to the supplementary materials.
\section{Experiments}
To evaluate the effectiveness of our method, we conduct comprehensive experiments on the tasks of cloth flattening and canonicalization (described below). To demonstrate the generalizability and robustness of our method, we evaluate in both the real world and in simulation on 5 different categories of garments. Through the experiments, we would like to answer the following questions:
\begin{enumerate}
    \item Does explicit occlusion reasoning improve the performance of cloth manipulation? How does it compare to methods that operate only on the visible points?
    \item Does test-time finetuning improve the quality of predicted mesh as well as the performance in cloth manipulation tasks?
    \item Can our method work on a physical robot?
\end{enumerate}

\subsection{Tasks}
\begin{figure}
     \centering
     \begin{subfigure}[b]{0.11\textwidth}
         \centering
         \includegraphics[width=\textwidth]{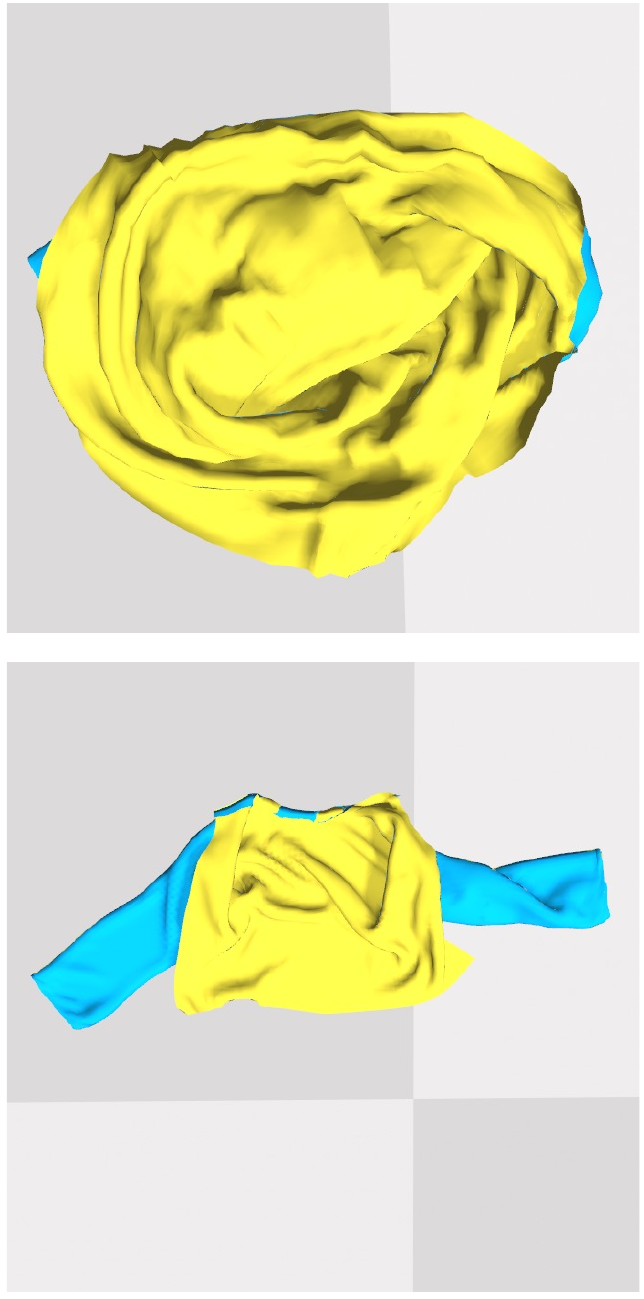}
         \caption{}
        \label{fig:why_canon}
     \end{subfigure}
     \hfill
     \begin{subfigure}[b]{0.32\textwidth}
         \centering
         \includegraphics[width=\textwidth]{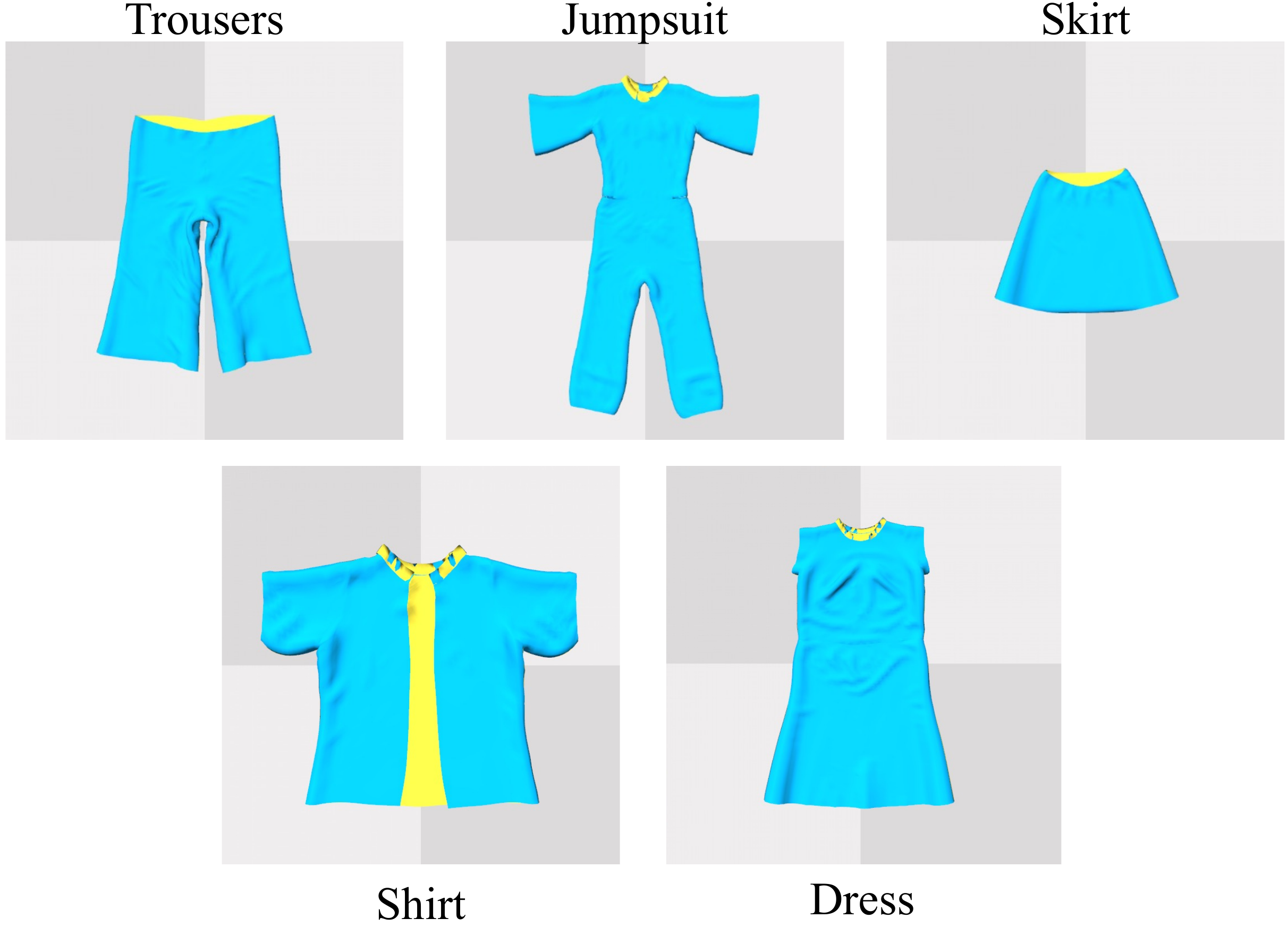}
         \caption{}
         \label{fig:canon_pose}
     \end{subfigure}
     \caption{Flattening (i.e. maximizing the covered area of the cloth) may not always give us a good starting point for folding. Left: Undesirable results of actions that optimize for the flattening task (maximizing the coverage). Right: Examplar goal poses for canonicalization of each category.}
     \vspace{-0.5cm}
\end{figure}

\noindent\textbf{Flattening}. Our goal is to flatten a crumpled cloth, that is, spreading it on the table.
Following prior works~\cite{lin2022learning}, we compute the coverage of the cloth as the objective for planning and evaluation. 

\noindent\textbf{Canonicalization}. Usually, flattening is the first step of a cloth manipulation pipeline~\cite{Li2018-sf,li2015folding,li2014real,mariolis2015pose,descriptors_fabrics_2021}, which makes the subsequent tasks such as folding easier. 
 However, for certain types of clothing, such as skirts or unbuttoned shirts, the flattening objective can produce undesirable results, as shown in Figure~\ref{fig:why_canon}, in which the robot maximizes the area covered by the cloth in a manner that is not conducive for downstream folding. 
 
Therefore, we also evaluate our method on a task that we call ``cloth canonicalization," where the goal is to manipulate the cloth and align it with the flattened canonical pose, as shown in Figure~\ref{fig:canon_pose}. To account for the ambiguity due to rotation and reflection symmetries, we define a set of symmetries for each type of cloth.  Using these symmetries, we define a canonical goal set of flattened poses $\mathcal{G}=\{G_i^{N\times3}\}_{i=1...A}$ for each cloth instance, where $A$ is the number of valid canonical poses. For example, for Trousers, we can rotate the canonical pose by 180 degrees to obtain another valid goal. The cost is computed as the minimum of the average pairwise distance to each of the possible canonical poses. Suppose that the current configuration of the cloth in the simulator is $V \in \mathbb{R}^{N\times 3}$, where $N$ is the number of vertices.   
Then the cost is computed as 
\begin{equation}
    Cost_{canon} = \min_{G_i\in \mathcal{G}} \frac{1}{N} \sum_{j=0}^{N-1} (g_j-v_j)^2
    \label{cost_canon}
\end{equation}
Note that $G_i$ and $V$ have the same number of vertices because both refer to the simulated cloth in different configurations.
In this work, we allow for our method to canonicalize the cloth without penalizing for errors with respect to a rigid transformations; this is because, for the task of cloth folding, the rotation and translation of the cloth is of lesser importance.  To evaluate this,
we first align the goal with the current state by computing an optimal rigid transformation using the Kabsch algorithm~\cite{arun1987least}, and then we compute the cost by Equation~\ref{cost_canon}.

\begin{figure*}[ht]
    \centering
    \includegraphics[width=1.0\textwidth]{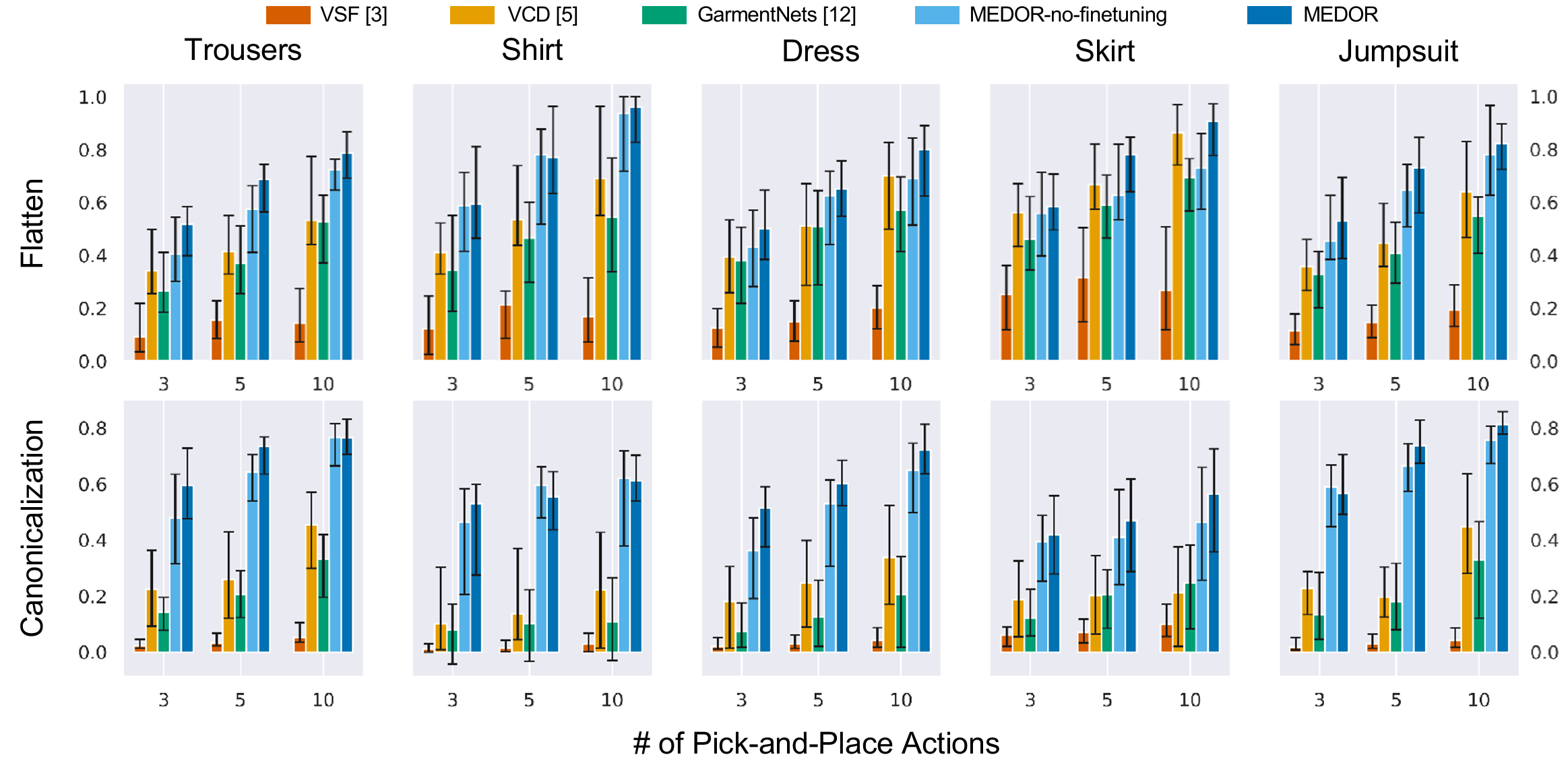}
    \caption{Normalized improvements on 2 tasks and 5 different categories of cloths. The height of the bar represents the median and the error bars show the 25 and 75 percentile of the performance.}
    \label{fig:main_results}
\end{figure*}

\subsection{Simulation Experiments}
\subsubsection{Baselines}
We compare our method to 4 baselines, including two state-of-the-art cloth manipulation methods:
\begin{itemize}
    \item \textbf{VisuoSpatial Foresight (VSF)}~\cite{fabric_vsf_2020}. This baseline learns a visual dynamics model in the RGB-D observation space. 
    It is trained on each category separately.
    \item \textbf{Visible Connectivity Graph (VCD)~\cite{lin2022learning}}. Similar to our method, VCD learns a particle-based dynamics model. However, unlike our method, VCD only operates on the visible points on the cloth, without explicitly reasoning about occlusions. An edge GNN is used to infer the mesh structure on the partial point cloud. To make a fair comparison with our method, the edge GNN is trained in a category-specific manner, while the dynamics model is only trained on Trousers (similar to our approach).
    \item \textbf{GarmentNets~\cite{chi2021garmentnets}}. In this baseline, we use the original implementation of GarmentNets  for mesh reconstruction, which processes a partial point cloud with PointNet++~\cite{qi2017pointnet++} and doesn't use relative prediction. For planning, we evaluate this baseline with the same mesh-based dynamics model as our method.
    \item \textbf{MEDOR-no-finetuning}. This is a variant of our method that removes the test-time finetuning step (Section~\ref{sec:Test-time finetuning}).
    \item \textbf{MEDOR}. This is our full method, which is essentially a modified version of GarmentNets with test-time finetuning.
\end{itemize}
For VSF and VCD, we provide the ground-truth RGB-D image and the ground-truth mesh (respectively) in the canonical pose to use for the reward computation. Note that our method does not have access to this information.
More details on the baselines can be found in the Supplement.


\subsubsection{Results}

For each task, we show the normalized improvement (NI) of each method, where 0 indicates no change from the initial state and 1 is the best possible performance. For flattening, we use the normalized improvement metric defined in previous work~\cite{lin2022learning}; for canonicalization, we compute the normalized improvement as $    \text{NI}_{canon} = \frac{cost_{init}-cost_{cur}}{cost_{init}}
    \label{eq:NI}$
where $cost_{init}$, $cost_{cur}$ are the costs of initial and current cloth configurations computed by Eq.~\ref{cost_canon}.
A trajectory is terminated after ten picks or when NI$>$0.95. 

For each category, we generate 40 initial configurations by random drops.
As shown in Figure~\ref{fig:main_results}, compared to methods without explicit occlusion reasoning (\emph{VCD}~\cite{lin2022learning} or \emph{VSF}~\cite{fabric_vsf_2020}), our methods MEDOR and MEDOR-no-finetuning both achieve competitive performance in all categories and both tasks. This shows the benefits of occlusion reasoning in cloth manipulation and the benefits of recovering the full configurations explicitly. 
Comparing \emph{MEDOR} against \emph{MEDOR-no-finetuning}, we can see the importance of test-time finetuning to adapt the mesh to better fit the observation.
Qualititative examples are shown in Figure~\ref{fig:smooth} as well as on the \href{https://sites.google.com/view/occlusion-reason/home}{website}, which show the differences in the mesh prediction before and after test-time finetuning.

For \emph{VSF} (orange) and \emph{VCD} (yellow), we see that the canonicalization task remains challenging even after being provided with the ground-truth RGB-D image or mesh for reward computation. This demonstrates the importance of planning with the completed cloth shape instead of planning only over the visible parts of the cloth.


We can also see that without our modifications, the original GarmentsNets~\cite{chi2021garmentnets} (green) has poor performance.
The variant of our method \emph{MEDOR-no-finetuning} includes the modifications to GarmentNets described in Section~\ref{method:coarse_estimate}: HRNet and Relative Prediction.
The huge performance gap between \emph{GarmentNets} and \emph{MEDOR-no-finetuning} demonstrates that these modifications lead to large performance benefits. 


    

\subsection{Ablations}
\label{sec:exp:ablation}

To further examine each component and design choice in the paper, we conduct the following ablations on both the flattening and canonicalization tasks.  Table~\ref{table:ablation} shows the normalized improvements averaged over 3, 5, and 10 picks, averaged over the flattening and canonlicalization tasks, and averaged over all 5 garment categories.


\textbf{Why is occlusion reasoning beneficial to cloth manipulation?}
Explicit occlusion reasoning improves the performance of our framework on cloth manipulation in two aspects: (1) Using the reconstructed mesh helps with the reward computation, and (2) Using the reconstructed mesh helps with the dynamics model. We differentiate these benefits in the following experiments:

(1)
In the 8th row \emph{Ours w/ Partial Reward}, we use only the visible portion of the cloth for reward computation, while the dynamics uses the full reconstruction (visible + occluded regions).
Compared to our full method, the performance drops by 29\%, showing the benefits of occlusion reasoning for the reward computation. 

(2) Using the reconstructed mesh improves the accuracy of the dynamics model: In VCD, the GNN dynamics model is trained and tested on the visible portion of the mesh. 
We compute the open loop rollout error of the VCD dynamics model on the visible mesh, and we likewise compute the rollout error of our method using the reconstructed mesh; we find that the rollout error of the VCD dynamics model (on the partial mesh) is 64.2\% higher than the rollout error on the reconstructed mesh. This demonstrates the importance of using the reconstructed mesh for accurate cloth dynamics. 

\begin{table}[t]
    \centering
    \begin{tabular}{c|c}
\toprule
\multirow{1}{*}{Method} & \multicolumn{1}{c}{Normalized}  \\ 
& Improvement\\
\hline
GarmentNets~\cite{chen2020learning}    &       0.320 $\pm$ 0.146  \\ 

No Mesh Reconstruction (VCD~\cite{lin2022learning}) & 0.391 $\pm$ 0.174 \\
No Finetuning and no Relative Prediction   & 0.560 $\pm$ 0.163       \\
No Finetuning   & 0.585 $\pm$ 0.171   \\ 
Joint Optimization & 0.614 $\pm$ 0.157 \\
No Consistency Loss    & 0.623 $\pm$ 0.148    \\
Replace GNN by GT Dynamics      &     0.631 $\pm$ 0.161    \\ 
Ours w/ Partial Reward & 0.462 $\pm$ 0.210 \\
\hline
\textbf{Ours (full method)}  & \textbf{0.651 $\pm$  0.138} \\
\hline
GT Mesh + Learned Dynamics    &     0.800 $\pm$ 0.096    \\ 
GT Mesh + GT Dynamics    &     0.870 $\pm$ 0.076
\end{tabular}
\caption{Ablation experiments. }
\label{table:ablation}
\end{table}
To further this analysis, we also perform an experiment in which we use the full reconstructed mesh for the dynamics model but use only the partial mesh for the reward computation (\emph{Ours w/ Partial Reward}).  If we compare the performance of this version to the 2nd row 
\emph{(No Mesh Reconstruction (VCD~\cite{lin2022learning}))}
we see the benefits of using the full reconstructed mesh for the dynamics model instead of the partial mesh (0.462 vs 0.391).



\textbf{How much does test-time finetuning help? Are both losses necessary?} When other components remain unchanged, we see that finetuning with only the Chamfer loss (\emph{No Consistency Loss}) already improves the performance over no finetuning (\emph{No Finetuning}) by 6.5\%.  Adding the Mapping Consistency Loss further boosts the performance from 0.623 (\emph{No Consistency Loss}) to 0.651 (\emph{Ours (full method)}); the combined improvement is 11.3\%.  Also, we find that without the 2-stage optimization scheme (\emph{Joint Optimization}), the mapping consistency loss hurts the performance (comparing \emph{No Consistency Loss} vs \emph{Joint Optimization}).

\textbf{ HRNet~\cite{sun2019deep} vs PointNet++~\cite{qi2017pointnet++}:} Looking at Table~\ref{table:ablation}, the only difference between the methods in the first and third row is the use of HRNet instead of PointNet++.  The huge difference in performance shows the benefits of the HRNet architecure for this task.

\textbf{Does Relative Prediction help?} Comparing the performance of \emph{No Finetuning and No Relative Prediction} with \emph{No Finetuning}, we see that this simple modification (described in Section~\ref{method:coarse_estimate}) improves performance.

\textbf{Do we need to learn the dynamics model instead of using the ground-truth dynamics model from the simulator?} 
Once we reconstruct the full cloth mesh, one option is to plan using the physics-based dynamics of Nvidia Flex simulator, as similarly done in \cite{chen2021ab}. This ablation is shown as \emph{Replace GNN by GT Dynamics} in Table~\ref{table:ablation}. We can see that this ablation yields a slight performance drop compared to our full method. We speculate that this is because the analytical dynamics model is more sensitive to mesh prediction errors than the learned dynamics model. As an additional point, planning with a simulator is 1.4 times slower than using a GNN dynamics model even after heavy parallelization (247 seconds vs. 103 seconds for 500 rollouts).

\textbf{Where are the remaining gaps in performance?}
As we can see in the last two rows, using the ground-truth mesh (instead of a learned mesh reconstruction) improves the performance by 23.1\%. We can improve performance another 7\% by also using the ground-truth dynamics model.  The remaining errors come from the sampling-based planner itself, which might fail to sample good actions.

\begin{figure}[b]
    \includegraphics[width=0.49\textwidth]{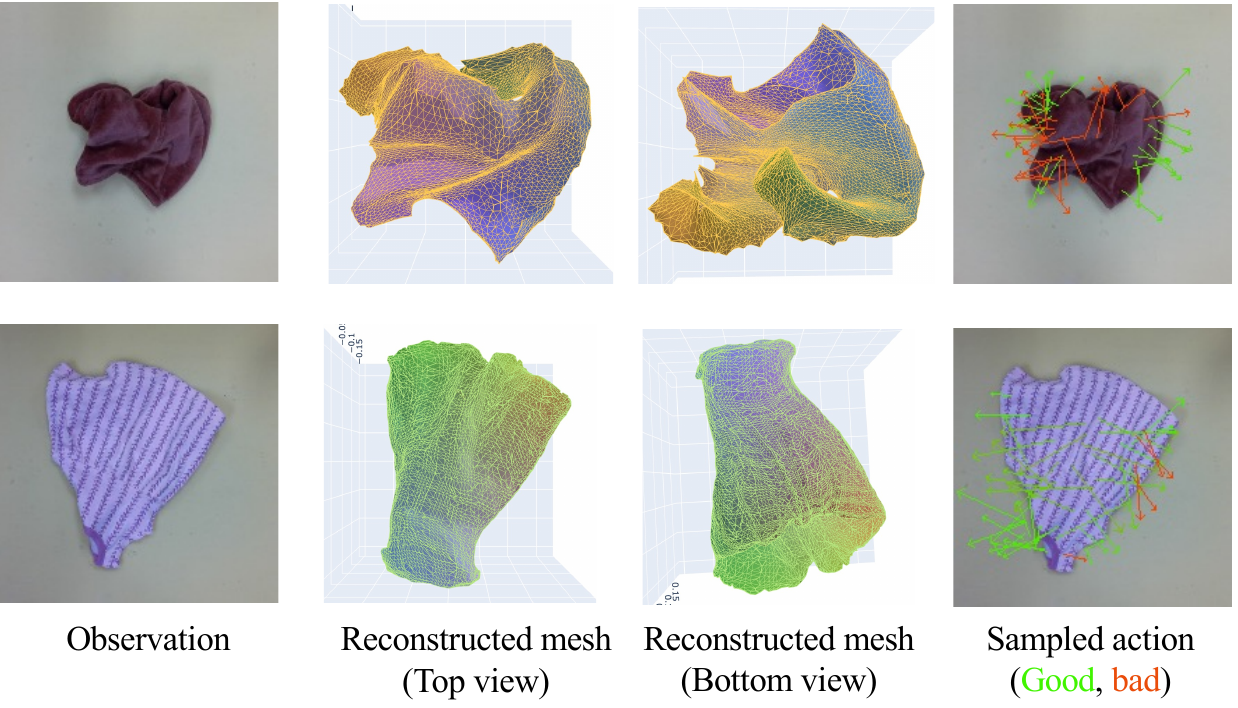}
    \caption{Example reconstruction results in the real-world, after test-time fine-tuning. In the 1st row, the trousers are successfully reconstructed, including the occluded legs and wrinkles on the surface. The planner (right column) is able to distinguish good actions from bad ones given the reconstructed mesh. In the 2nd row, our model failed to capture the left-bottom corner which is folded under the visible layer. As such, the planner failed to choose actions to reveal the occluded part.}
    \label{fig:real_recon}
\end{figure}



\subsection{Physical Experiments}
\begin{figure*}[h!]
    \centering
    \includegraphics[width=\textwidth]{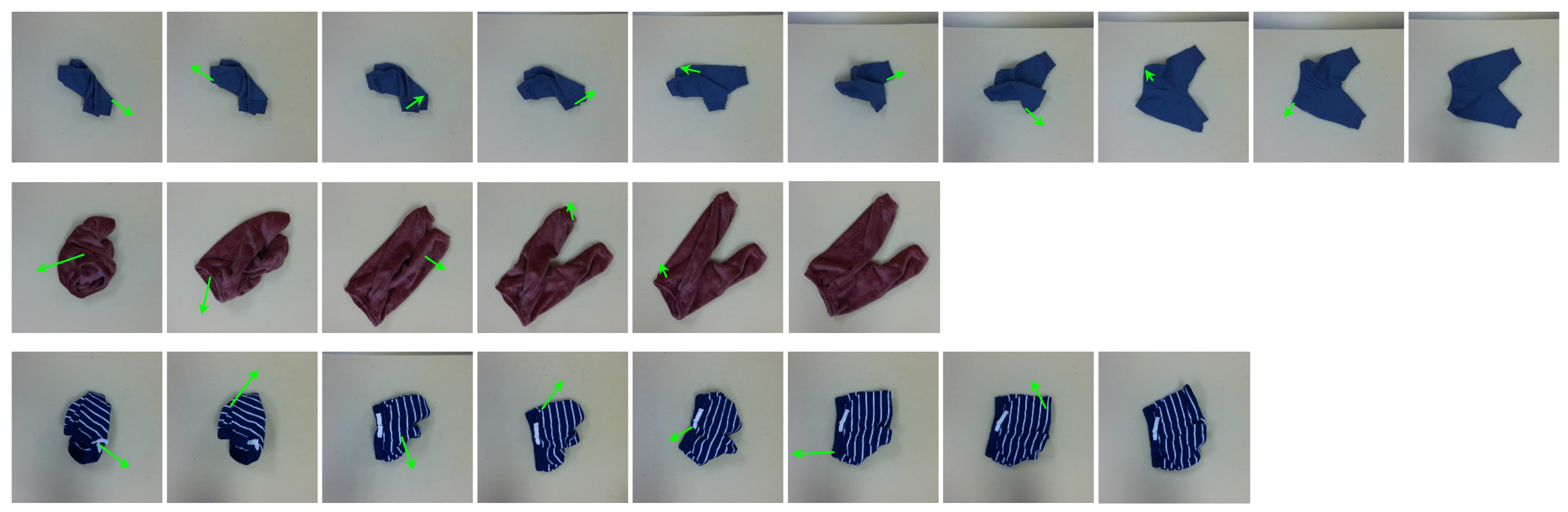}
    \caption{We evaluate our method of 5 pieces of clothing. Please refer to our \href{https://sites.google.com/view/occlusion-reason/home}{website} for videos.}
    \label{fig:qualitative_robot}
    \vspace{-0.5cm}
\end{figure*}

\begin{figure}
    \includegraphics[width=0.49\textwidth]{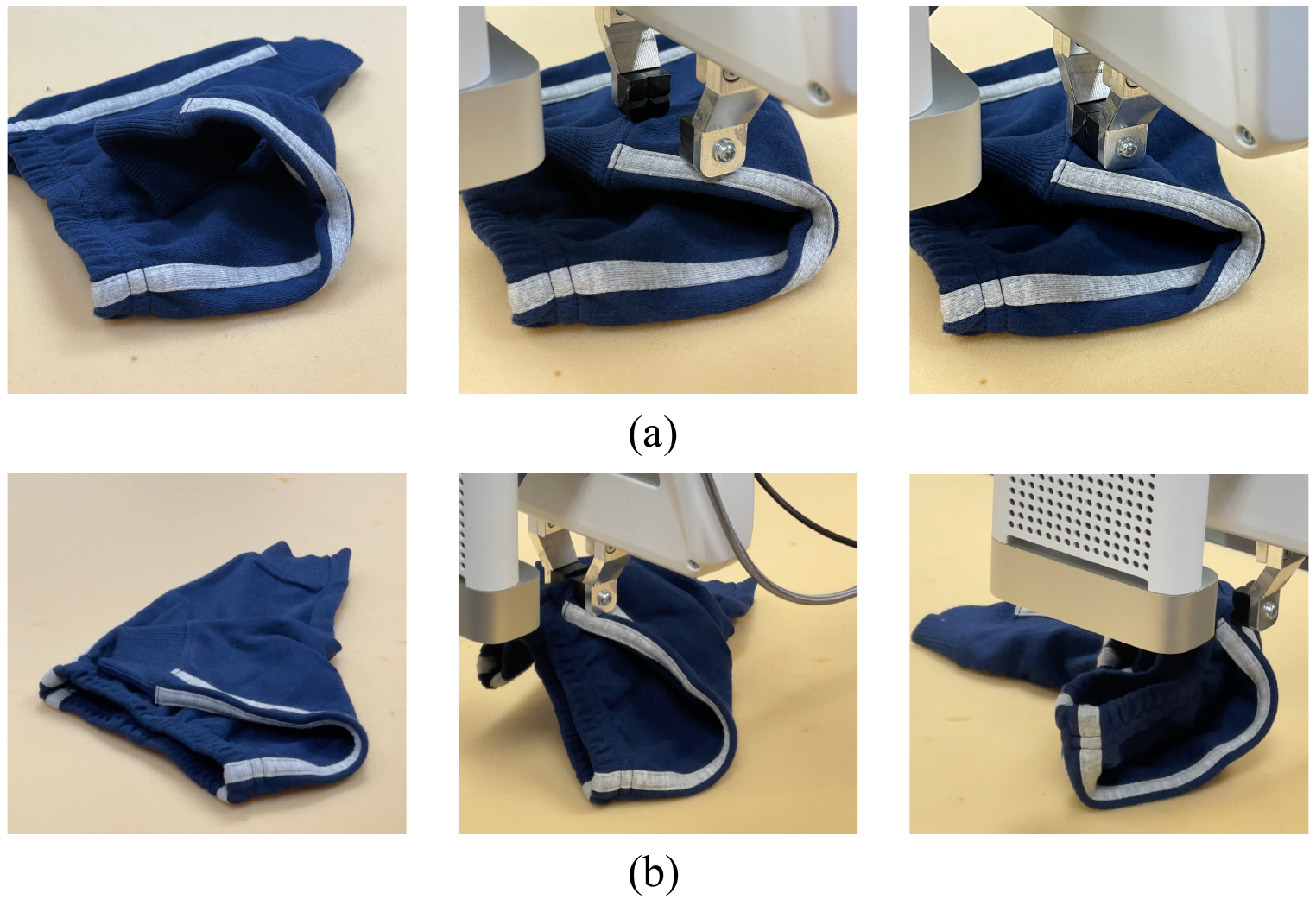}
    \caption{ Grasping failures: In (a), the cloth is deformed when the gripper moves down, resulting in missed grasping. In (b), the robot is supposed to grasp the upper layer and unfold it, but it mistakenly grasps the bottom layer as well.}
    \label{fig:grap_failure}
    \vspace{-0.5cm}
\end{figure}

We also evaluate our method in the real world by deploying it on a 7-DOF Franka Emika Panda robot. We mount an Azure Kinect depth sensor on the end-effector of the robot. When taking the depth image, the end-effector will move to be centered above the cloth. 
To obtain a valid plan for pick and place actions, we use MoveIt!~\cite{chitta2012moveit}.

We evaluate our method on Trousers (3 instances) and Dress (2 instances) on the cloth flattening task.  For each article of clothing, we run 5 trajectories with at most 10 pick-and-place actions each. The trajectories will be terminated if the normalized improvements exceed 95\%. We obtain random configurations by performing a random drop three times. We compare our method with two baselines. \emph{Random} is a heuristic policy that performs random pick-and-place actions. The picked points are biased towards contour of the cloth and the place points are always outside the cloth region. VCD~\cite{lin2022learning} is a prior method that plans with a partial point cloud instead of reconstructing the full mesh.
As shown in  Fig.~\ref{fig:qualitative_robot}, our model can efficiently smooth the clothes by only a few pick-and-place actions. The performance of our method compared to the baselines is shown in Table~\ref{table:real_result}. 
\begin{table}
    \centering
    \begin{tabular}{c|c|c|c}
\toprule
& Random & VCD~\cite{lin2022learning} & MEDOR (Ours)\\
\midrule
Trousers & 0.044 & 0.562 & \textbf{0.647}\\
\midrule
Dress & 0.036 & 0.361 & \textbf{0.468}
\end{tabular}
\caption{Results of physical robot experiments. }
\vspace{-0.5cm}
\label{table:real_result}
\end{table}

We do observe a gap between the performance in simulation and in the  real-world. The first source of error is the reconstruction error. Since the model is trained in simulation, it suffers from a distribution shift resulting from the difference in modeling the cloth physics and material properties, e.g., stiffness, thickness. In Figure~\ref{fig:real_recon}, we show a successful and  a failed reconstruction result.  Another source of error comes from the grasp execution, which mostly occur when multiple layers of cloth are stacked together. There are two main failure modes: (1) The robot gripper deforms the cloth, causing a failed grasp; (2) The robot grasps the wrong number of cloth layers. Figure~\ref{fig:grap_failure} illustrates the two cases.

\section{Conclusions}
We introduce a cloth manipulation system that explicitly reasons about the occluded regions of cloth. 
At test-time, we optimize the predicted mesh with self-supervised losses. Then we use a learned mesh-based dynamics model to plan over the predicted mesh and find optimal actions for cloth manipulation. We compare against state-of-the-art cloth manipulation methods that do not account for partial observability and show significant improvements from explicit occlusion reasoning and test-time fine-tuning. We also demonstrate the efficacy of our method in real world experiments.

\subsection*{Acknowledgments}
\noindent
This work was supported by LG Electronics, National Science Foundation (NSF) CAREER Award (IIS-2046491) and NSF Smart and Autonomous Systems Program (IIS-1849154).

\newpage
\bibliographystyle{unsrt}
\bibliography{main}

\newpage


\section{Experiments Setup}

\subsection{Simulation Setup}

\begin{table}[b]\centering
\begin{tabular}{@{}lp{30mm}}

\toprule
Simulation parameters & Value\\
\midrule
\hspace{5mm}Camera view &  Top-down\\
\hspace{5mm}Camera position & [0, 0.65 m, 0] \\
\hspace{5mm}Field of view & 90\\
\hspace{5mm}Picker radius & 0.01 m\\
\hspace{5mm}Picker threshold & 0.00625 m\\
\hspace{5mm}dt & 0.01 second\\

\hspace{5mm}Damping & 1\\
\hspace{5mm}Dynamic friction & 1.2\\
\hspace{5mm}Particle friction & 1\\
\hspace{5mm}Stiffness (strech, bend, shear) & [1.2, 0.6, 1]\\
\hspace{5mm}Mass & 0.0003\\
\hspace{5mm}Particle radius & 0.005\\
\hspace{5mm}Gravity & -9.8\\

\bottomrule
\end{tabular}
\caption{Hyper-parameters of softgym.}
\label{tab:sim_params}
\end{table}

\begin{table}[b]
    \centering
    \begin{tabular}{c|c|c|c|c}
    \toprule
    Trousers & Shirt & Dress & Skirt & Jumpsuit\\
    \midrule
    2 & 1 & 2 & 12& 2
    \end{tabular}
    \caption{Order of rotation symmetry of different types of cloth. The order of rotation symmetry is the times the shape fits onto itself when rotating for 360 degrees. An order of 2 means that the shape remains unchanged when rotating for 180 or 360 degrees.  We use it to construct goal sets for cloth canonicalization task.}
    \label{tab:canon_symmetry}
\end{table}
\begin{table*}[h]
    \centering
         \tabcolsep=0.08cm
         \small 
\begin{tabular}{l|ccccc}
\toprule
      & Trousers & Shirt & Dress & Skirt& Jumpsuit\\ \hline
No. of Meshes     & 1691     & 1111    & 2037         & 468    & 2279         \\

Avg. No. of Vertices    & $7050\pm1954 $ & $5308\pm996 $  & $8030\pm2159$ & $4643\pm1225$ & $10686\pm 2572$ \\
Avg. No. of Vertices (downsampled) & $278\pm87$  & $214\pm59$  & $291\pm99$  & $190\pm80$  & $215\pm58$   \\
Rescaling Factor & 0.42 & 0.36 & 0.29 & 0.28 & 0.28 \\
Avg. X (m)  & $0.29\pm0.03$  & $0.32\pm0.10$   & $0.24\pm0.07$  & $0.20\pm0.05 $   & $0.19\pm0.08$         \\
Avg. Y (m)  & $0.12\pm0.02$  & $0.11\pm0.02$   & $0.15\pm0.05$  & $ 0.15\pm0.05 $   & $0.09\pm0.02$        \\
Avg. Z (m)  & $0.27\pm0.08$  & $0.19\pm0.04$   & $0.31\pm0.06$  & $0.18\pm0.05 $   & $0.31\pm0.06$        \\
\bottomrule
\end{tabular}
\caption{The statistics of the CLOTH3D dataset~\cite{bertiche2020cloth3d} after pre-processing.}
\label{table:stats_cloth3d}
\end{table*}

We conduct all simulation experiments in Softgym~\cite{corl2020softgym}, a simulation environment for deformable objects built on the particle-based simulator, Nvidia Flex. We model a flying gripper as a spherical picker that can move freely in the 3D space. When a cloth particle is "picked", it will move rigidly with the gripper. The simulation parameters of softgym can be found in Table.~\ref{tab:sim_params}. We obtained the 3D cloth models from CLOTH3D dataset~\cite{bertiche2020cloth3d}. Considering the physical experiments, we rescale the cloth models so that they could fit in the workspace of our real robot. Also, to make the GNN dynamics computationally feasible, we create a downsampled version of the cloth models. During data collection, we still use the original dense mesh in softgym, but register the downsampled mesh onto the dense mesh by finding the nearest neighbor of each vertex. Therefore, we obtain the trajectory of downsampled mesh and use that for dynamics learning. The detailed specifications of the preprocessed CLOTH3D dataset can be found in Table.~\ref{table:stats_cloth3d}. 
We use a top-down camera and it's placed at a fixed height of 0.65 m. The valid workspace is 0.53 m $\times$ 0.53 m.

\textbf{Flattening} For flattening task, the goal is to maximize the coverage of the cloth in the current configuration.  To compute the coverage, we treat each node on the graph as a sphere with a radius of 0.005 and compute the covered area when projected to the ground plane.

\textbf{Canonicalization} In CLOTH3D, the canonical pose of the cloth is defined to be the pose of cloth when wore by a T-pose human. However, since we are considering cloth manipulation on a planar surface, it is not achievable and cannot be used as the goal pose directly. Therefore, we use gravity (10 times the gravity on Earth) to obtain a flattened version of the canonical pose. Another thing that needs to be handled is the ambiguity caused by reflection symmetry and rotation symmetry. For example, trousers are approximately 180$^\circ$ rotation symmetry, which means that the shapes before and after rotating around the center axis for 180$^\circ$ are the same. Therefore, during the evaluation of cloth canonicalization, we define a canonical goal set $\mathcal{G}=\{G_i^{N\times3}\}_{i=1...A}$, for each type of cloth. $A$ is the number of valid canonical poses. The cost is computed as the minimum of average pairwise distance to each of the possible canonical poses. Suppose that the current configuration of the cloth is $V^{N\times 3}$, where $N$ is the number of vertices, $g_j$ and $v_j$ is the $j$-th vertex of the mesh, the cost is computed as 
\begin{equation}
    Cost_{canon} = \min_{G_i\in \mathcal{G}} \sum_{j=n}^{N} \frac{(g_j-v_j)^2}{N}
    \label{cost_canon}
\end{equation}

The rotation symmetry of each category is described in Table.~\ref{tab:canon_symmetry}. It should be noted that since we maintain a discrete set of plausible goal pose, although skirt has infinite order of rotation symmetry, we cannot iterate over all possible cases. Instead, we make an approximation that the order of rotation symmetry of skirt is 12. The order of rotation symmetry is the times the shape fits onto itself when rotating for 360 degrees. An order of 2 means that the shape remains unchanged when rotating for 180 or 360 degrees.

\section{Additional results}

\begin{table*}[t]
    \centering
    \scriptsize
    \begin{tabular}{c|c|ccc|ccc}
    \toprule
    & Task&\multicolumn{3}{c|}{Flattening} & \multicolumn{3}{c}{Canonicalization} \\
    & Number of Pick-and-Place & 1 & 2 & 3 & 1 & 2 & 3\\ 
    \bottomrule
     
    \multirow{5}{*}{Trousers} 
    &  VSF~\cite{fabric_vsf_2020}  & $0.09\pm0.13$   & $0.15\pm0.07$   & $0.14\pm0.13$ & $0.02\pm0.02$   & $0.03\pm0.04$   & $0.05\pm0.05$ \\
    &  VCD~\cite{lin2022learning}  & $0.34\pm0.15$   & $0.41\pm0.14$   & $0.53\pm0.24$ & $0.22\pm0.14$   & $0.26\pm0.17$   & $0.46\pm0.16$ \\
    & GarmentNets~\cite{chi2021garmentnets}      & $0.27\pm0.15$   & $0.37\pm0.14$   & $0.53\pm0.16$ & $0.14\pm0.06$   & $0.21\pm0.09$   & $0.33\pm0.14$\\
    & MEDOR (no fine-tuning)   & $0.41\pm0.14$   & $0.57\pm0.16$   & $0.72\pm0.08$ & $0.48\pm0.16$   & $0.64\pm0.10$   & $0.77\pm0.10$ \\
    & MEDOR  & $\mathbf{0.52\pm0.12}$   & $\mathbf{0.69\pm0.12}$   & $\mathbf{0.79\pm0.10}$  & $\mathbf{0.59\pm0.13}$   & $\mathbf{0.73\pm0.10}$   & $\mathbf{0.77\pm0.07}$ \\
    \bottomrule
    
    \multirow{5}{*}{Shirt} 
    & VSF~\cite{fabric_vsf_2020}   & $0.12\pm0.07$   & $0.15\pm0.08$   & $0.20\pm0.09$ & $0.01\pm0.02$   & $0.01\pm0.03$   & $0.03\pm0.04$ \\
    & VCD~\cite{lin2022learning}    & $0.39\pm0.14$   & $0.51\pm0.23$   & $0.70\pm0.20$ & $0.10\pm0.20$   & $0.14\pm0.23$   & $0.22\pm0.21$ \\
    & GarmentNets~\cite{chi2021garmentnets}  & $0.34\pm0.21$   & $0.47\pm0.17$  & $0.55\pm0.22$ & $0.08\pm0.12$   & $0.10\pm0.14$   & $0.11\pm0.16$ \\
    & MEDOR (no fine-tuning)  & $\mathbf{0.59\pm0.17}$   & $\mathbf{0.78\pm0.26}$ & $0.94\pm0.22$ & $0.46\pm0.26$   & $\mathbf{0.60\pm0.12}$   & $\mathbf{0.62\pm0.24}$ \\
    & MEDOR & $0.59\pm0.22$   & $0.77\pm0.19$   & $\mathbf{0.96\pm0.13}$ & $\mathbf{0.53\pm0.26}$   & $0.56\pm0.12$   & $0.61\pm0.09$\\
    \bottomrule
    
    \multirow{5}{*}{Dress} 
    & VSF~\cite{fabric_vsf_2020}            & $0.12\pm0.07$   & $0.15\pm0.08$   & $0.20\pm0.09$ & $0.02\pm0.03$   & $0.03\pm0.03$   & $0.04\pm0.04$ \\
    & VCD~\cite{lin2022learning}            & $0.39\pm0.14$   & $0.51\pm0.23$   & $0.70\pm0.20$ & $0.18\pm0.17$   & 
    $0.25\pm0.16$   & $0.34\pm0.19$ \\
    & GarmentNets~\cite{chi2021garmentnets}      & $0.38\pm0.16$   & $0.51\pm0.22$   & $0.57\pm0.16$ & $0.07\pm0.10$   & $0.13\pm0.13$   & $0.21\pm0.19$ \\
    & MEDOR (no fine-tuning)                & $0.43\pm0.15$   & $0.63\pm0.19$   & $0.69\pm0.18$ & $0.36\pm0.17$   & $0.53\pm0.22$   & $0.65\pm0.15$  \\ 
    & MEDOR                                 & $\mathbf{0.50\pm0.14}$   & $\mathbf{0.65\pm0.11}$   & $\mathbf{0.80\pm0.17}$  & $\mathbf{0.51\pm0.14}$   & $\mathbf{0.60\pm0.08}$   & $\mathbf{0.72\pm0.09}$\\

    \bottomrule
    \multirow{5}{*}{Skirt} 
    & VSF~\cite{fabric_vsf_2020}            & $0.25\pm0.13$   & $0.32\pm0.19$   & $0.27\pm0.24$  & $0.06\pm0.04$  & $0.07\pm0.05$   & $0.10\pm0.07$  \\
    & VCD~\cite{lin2022learning}            & $0.56\pm0.13$   & $0.67\pm0.15$   & $0.87\pm0.12$ & $0.19\pm0.14$  & $0.20\pm0.14$   & $0.21\pm0.19$ \\
    & GarmentNets~\cite{chi2021garmentnets}      & $0.46\pm0.16$   & $0.59\pm0.13$   & $0.70\pm0.13$ & $0.12\pm0.10$  & $0.20\pm0.12$   & $0.25\pm0.16$ \\
    & MEDOR (no fine-tuning)                & $0.56\pm0.16$   & $0.63\pm0.19$   & $0.73\pm0.16$ & $0.39\pm0.14$  & $0.41\pm0.17$  & $0.46\pm0.21$ \\
    & MEDOR                                 & $\mathbf{0.58\pm0.12}$   & $\mathbf{0.78\pm0.14}$   & $\mathbf{0.91\pm0.13}$  & $\mathbf{0.42\pm0.14}$   & $\mathbf{0.47\pm0.18}$   & $\mathbf{0.56\pm0.21}$  \\

    \bottomrule
    \multirow{5}{*}{Jumpsuit} 
    & VSF~\cite{fabric_vsf_2020}            & $0.12\pm0.06$   & $0.15\pm0.07$   & $0.19\pm0.10$ & $0.02\pm0.03$ & $0.03\pm0.03$ & $0.04\pm0.05$ \\
    & VCD~\cite{lin2022learning}            & $0.36\pm0.10$   & $0.45\pm0.15$   & $0.64\pm0.19$ & $0.23\pm0.09$   & $0.19\pm0.11$   & $0.45\pm0.19$ \\
    & GarmentNets~\cite{chi2021garmentnets}      & $0.33\pm0.12$   & $0.41\pm0.12$   & $0.55\pm0.14$ & $0.13\pm0.15$   & $0.18\pm0.14$   & $0.33\pm0.21$ \\
    & MEDOR (no fine-tuning)                & $0.45\pm0.17$   & $0.65\pm0.14$   & $0.78\pm0.19$ & $\mathbf{0.59\pm0.14}$   & $0.67\pm0.09$   & $0.76\pm0.08$\\
    & MEDOR                                 & $\mathbf{0.53\pm0.17}$   & $\mathbf{0.73\pm0.17}$   & $\mathbf{0.82\pm0.10}$ & $0.57\pm0.14$   & $\mathbf{0.74\pm0.09}$   & $\mathbf{0.81\pm0.05}$ \\

    \bottomrule
    \end{tabular}
    \caption{Normalized Improvement (NI) of cloth flattening and cloth canonicalization, for varying numbers of allowed pick and place actions.}
    \label{tab:simulation_performance}
\end{table*}
\begin{figure*}[ht!]
    \centering
    \includegraphics[width=\textwidth]{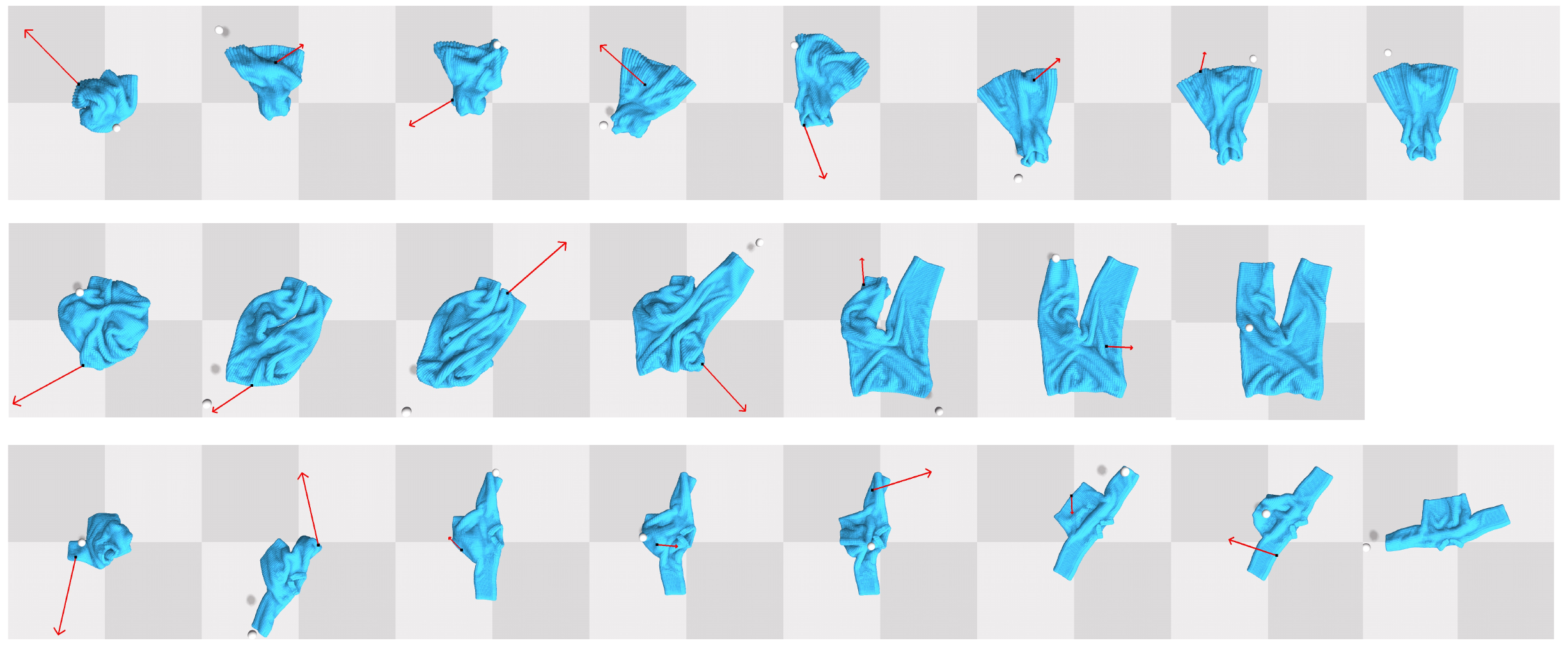}
    \caption{Examplar trajectories of canonicalization task in simulation. As we can see, our method is able to quickly unfold the cloths from extremely crumpled configurations in a few steps. }
    \label{fig:qualitative_sim}
\end{figure*}

\subsection{Simulation Experiments}

The full results of simulation are shown in the Table.~\ref{tab:simulation_performance}, best performance within each category is bolded.
We also show several trajectories of cloth canonicalization in Fig.~\ref{fig:qualitative_sim}.
\begin{figure*}[ht]
    \centering
    \includegraphics[width=0.9\textwidth]{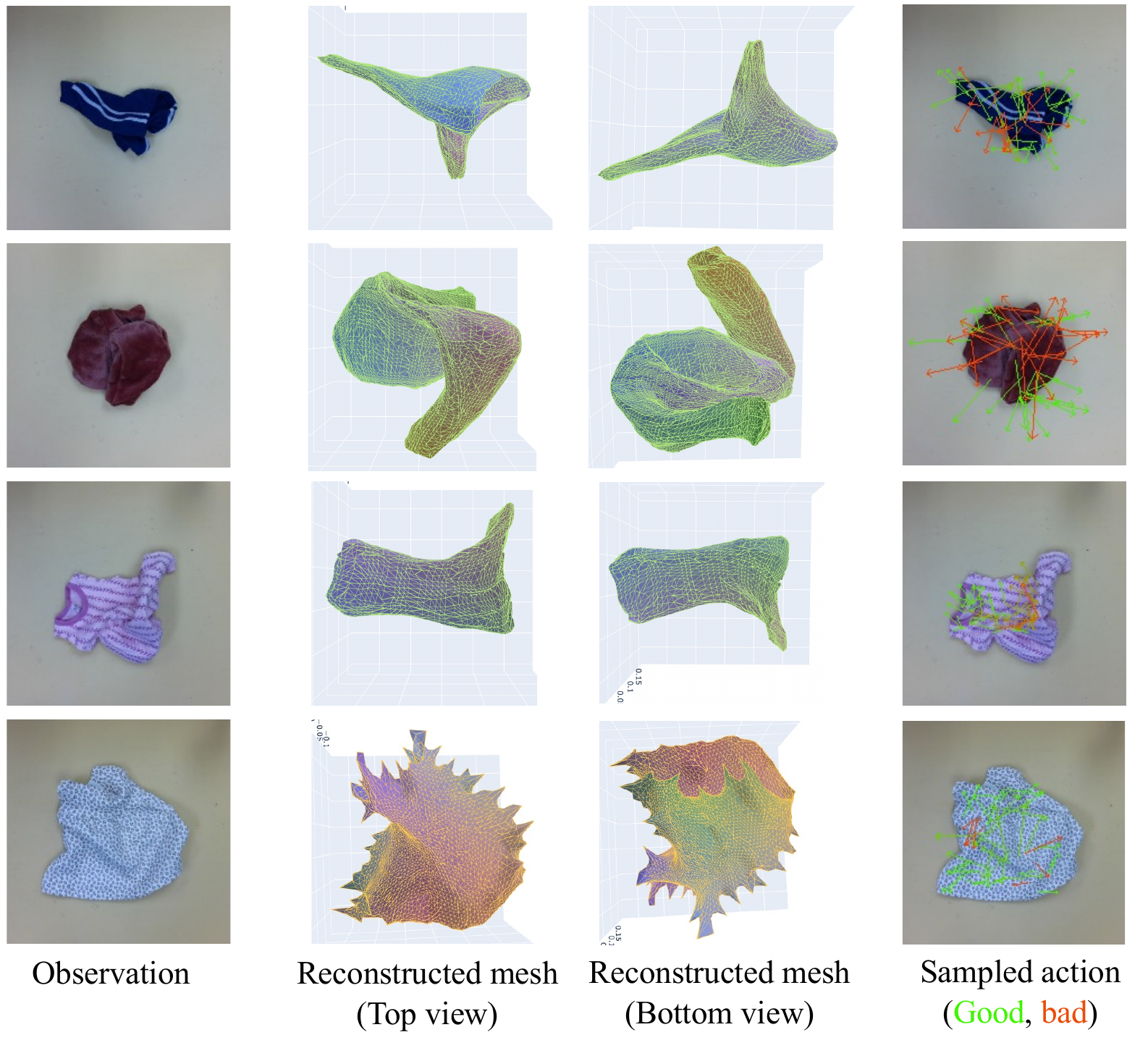}
    \caption{Reconstruction results in real world.}
    \label{fig:real_rec}
\end{figure*}
\begin{figure*}[ht]
    \centering
    \includegraphics[width=0.9\textwidth]{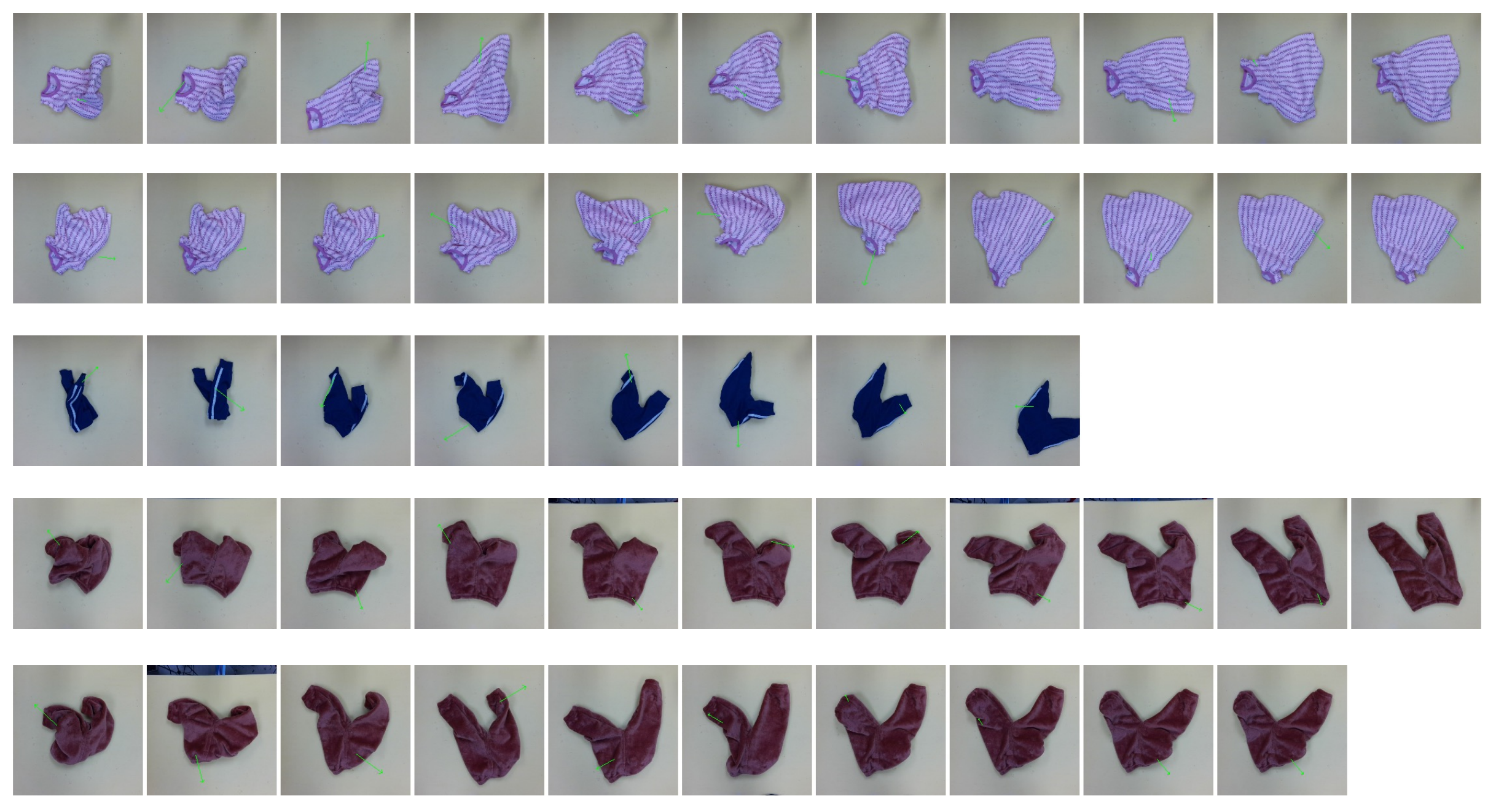}
    \caption{Rollout of cloth flattening in real world.}
    \label{fig:real_rollouts}
\end{figure*}

\subsection{Physical Experiments}
Some additional results of physical experiments are shown in Fig.~\ref{fig:real_rec} and Fig.~\ref{fig:real_rollouts}.

\section{Implementation details of MEDOR}
\label{sup:implement}
\subsection{GarmentNets-style Mesh Reconstruction Model}
As described in the main paper, we employ a GarmentNets-style model to reconstruct the mesh from depth image. GarmentNets formulates the pose estimation problem of clothes as a shape completion task in the canonical space. By doing so, the model learns meaningful correspondence between There are several advantages about GarmentNets: 1) it reconstructs the occluded part of the clothes due to self-occlusion; 2) it estimates the correspondence between clothes in canonical space and observation space; 3) it estimates the pose of the clothes (per-vertex location). 

\subsubsection{Model Architecture}
We build our mesh reconstruction model based on GarmentNets~\cite{chi2021garmentnets} with some decisive modifications to make it fit in our setup. Here, we give a brief description of the reconstruction pipeline and the modifications we made. For details, please refer to GarmentNets~\cite{chi2021garmentnets}.

\textbf{Canonicalization}  Given the observation at current state, we first use \emph{canonicalization model} to map it to canonical space by conducting pixel-wise canonical coordinate prediction. We don't pick up the clothes for state estimation, because it may disrupt the configuration of partially folded or almost smooth clothes. We use depth images captured by a top-down camera as the observation and High-Resolution Network (HRNet-32)~\cite{sun2019deep} as the backbone of canonicalization model. HRNet is a convolutional architecture that specializes in producing high-resolution and spatially precise representations. It uses smaller convolutional kernels and avoids overly downsampling the feature map. This is critical for our task because our model needs to infer the structure of crumpled clothes by subtle changes at the contour of different layers.  The architecture change improves the performance on cloth smoothing and canonicalization tasks for 75\%. Following GarmentNets~\cite{chi2021garmentnets}, we formulate the prediction problem as a classification task by dividing each axis into 64 bins and use Cross-entropy loss for training.
\textbf{Feature Scattering} Given the predicted canonical coordinate, we scattered the features of each pixel to corresponding locations in the canonical space. The aggregated feature volume is further transformed by a 3D UNet~\cite{cciccek20163d}, which is trained by shape completion and flow prediction.

\textbf{Shape Completion} 
Before estimating the pose of occluded surface, we first perform volumetric shape completion in the canonical space, which helps capture the shared structure within the same category. Since the structure of clothes are thin and non-watertight, GarmentNets~\cite{chi2021garmentnets} proposes to use Winding Number Field~\cite{jacobson2013robust} as the shape representation. The shape completion network is instantiated as an implicit network. It takes the dense feature produced by 3D UNet and a canonical coordinate and output the winding number field in that coordinate. 

\textbf{Predicting pose in the observation space}
After we complete the shape of clothes in the canonical space, we estimate the pose of the clothes in observation space. We cast it as a 3D flow prediction problem by predicting per-vertex flow that transforms the clothes from canonical space to observation space.
\begin{equation}
    \tilde{x}_i^o = \tilde{x}_i^c + \tilde{f}_i
\end{equation}

\subsubsection{Training and Testing Details}. Now we describe how we train and test the model.

\textbf{Dataset collection}. Our model is category-specific, so we collect a dataset for each category separately in Softgym~\cite{corl2020softgym}. We obtained 3D clothes models from CLOTH3D dataset~\cite{bertiche2020cloth3d}.

Each dataset contains 4,000 trajectories of length 5 for training and 400 for testing.  The At the beginning of each trajectory, we randomly sample a clothes mesh and initialize it by random drop or flattened pose with equal probability. Then we disrupt the clothes with random pick-and-place actions. Random actions are biased towards picking corners (obtained by Harris corner detection\cite{Harris1988ACC}) 90\% of the time, otherwise it is sampled uniformly on the clothes. The distance between the pick point and the place point is uniformly sampled between [25, 150] pixels.

\textbf{Training details} The model is trained in two-stage. In the first stage, we train the canonicalization network with Cross-entropy loss till convergence. In the second stage, we freeze the canonicalization network and train the rest of the models for shape completion and flow prediction by Mean-square error.

\textbf{Inference details} At test-time, given a depth image, we first use canonicalization network and 3D UNet to obtain dense feature volume in the canonical space. Then we discretize the canonical space into 128x128x128 grid and evaluate shape completion network at every cell. To retrieve the mesh, we compute the Gaussian derivatives for the predicted winding number field and run Marching Cube algorithm~\cite{lorensen1987marching}.

\begin{table}[t]\centering
\begin{tabular}{@{}lp{30mm}}
\toprule
Model parameter & Value\\
\midrule
\textit{Canonicalization Network}\\
\hspace{5mm}Backbone & HRNet-32 \\
\hspace{5mm}Dimension of output feature & 489 \\
\textit{3D CNN}\\
\hspace{5mm}Backbone & 3D Unet \\
\hspace{5mm}Level & 4 \\
\hspace{5mm}Feature maps & 32\\
\hspace{5mm}Dimension of output feature & 128 \\
\textit{Implicit Shape Completion Network}\\
\hspace{5mm}Backbone & MLP\\
\hspace{5mm}Number of hidden layers & 3 \\
\hspace{5mm}Size of hidden layers & 512 \\
\textit{Implicit Shape Completion Network}\\
\hspace{5mm}Backbone & MLP\\
\hspace{5mm}Number of hidden layers & 3 \\
\hspace{5mm}Size of hidden layers & 512 \\
\toprule
Training parameters & Value\\
\midrule
\hspace{5mm}Optimizer & Adam \\
\hspace{5mm}Learning rate & 0.0001 \\
\hspace{5mm}Gaussian noise std & 0.005 \\
\hspace{5mm}Random rotation & [-180, 180] \\

\end{tabular}
\caption{Hyper-parameters of mesh reconstruction model}
\label{tab:hyper_params}
\end{table}

\subsection{Dynamics Model}
Similar to VCD~\cite{lin2022learning}, we use a learned GNN-based dynamics model proposed in GNS~\cite{sanchez2020learning}. The difference between VCD and ours is that we don't have a GNN edge model because edges are estimated by our mesh reconstruction model. The original mesh models in CLOTH3D~\cite{bertiche2020cloth3d} ~(see Table.~\ref{table:stats_cloth3d}) are too dense that the rollout becomes computationally infeasible. Therefore, we downsample the mesh by using Vertex Clustering~\cite{low1997model} with a voxel size of 0.025m. For the complete list of hyperparameters of the GNN dynamics model, please refer to Table~\ref{tab:dyn_params}.

\begin{table*}[h!]\centering
\begin{tabular}{@{}lp{40mm}}
\toprule
Model parameter & Value\\
\midrule
\textit{Encoder(same for both node encoder and edge encoder)} & \\
\hspace{5mm}Number of hidden layers & 3 \\
\hspace{5mm}Size of hidden layers & 128 \\

\textit{Processor} & \\
\hspace{5mm}Number of message passing steps & 10 \\
\hspace{5mm}Number of hidden layers in each edge/node update MLP  & 3 \\
\hspace{5mm}Size of hidden layers & 128 \\
\textit{Decoder} & \\
\hspace{5mm}Number of hidden layers & 3 \\
\hspace{5mm}Size of hidden layers & 128 \\
\toprule
Training parameters & Value\\
\midrule
\hspace{5mm}Number of trajectories & 5000\\
\hspace{5mm}Learning rate & 0.0001\\
\hspace{5mm}Batch size & 16 \\
\hspace{5mm}Training epoch & 120\\
\hspace{5mm}Optimizer & Adam\\
\hspace{5mm}Beta1 & 0.9\\
\hspace{5mm}Beta2 & 0.999\\
\hspace{5mm}Weight decay & 0\\
\toprule
Others & Value\\
\midrule
\hspace{5mm}dt  & 0.05 second \\
\hspace{5mm}Particle radius & 0.005 m \\
\hspace{5mm}Vertex clustering voxel size & 0.025 m \\ 
\hspace{5mm}Neighbor radius $R$ & 0.036 m\\
\bottomrule
\end{tabular}
\caption{Hyper-parameters of GNN dynamics model.}
\label{tab:dyn_params}
\end{table*}

\subsection{Planning}
\label{sup:medor:plan}
The planning algorithm is outlined in Algorithm~\ref{algo:planning}. We plan in the space of pick-and-place primitive with horizon equal to 1. To simulate the effect of each pick-and-place, we divide them into low-level actions and roll out by the dynamics model in parallel. Following \cite{lin2022learning}, the action is encoded into the input mesh by directly modifying the position and picked point, and the displacement will be propagated to the rest of mesh during message passing. 

\textbf{Action sampling during planning} For both cloth flattening and canonicalization, we bias the actions sampling toward the contour of the cloth. More specifically, we first obtain the bounding box of the cloth in current observation and expand it by 30 pixels in each direction. Then we randomly sample picked points within the the bounding box region. For points that are not on the cloth, we map them to the nearest point on the cloth. The place direction is uniformly sampled in all directions and place distance is sampled uniformly from [0.05, 0.2]. A dummy action which corresponds to "no action" is added to the list of candidate actions. We sample 500 pick-n-place actions at each timestep.

\textbf{Reward computation} For flattening, we treat each mesh vertex as a sphere of radius 0.01, and the total reward is the covered area of the projection of all vertices to the ground plane. For canonicalization, we rotate the predicted canonical pose according to the predefined rotation symmetry (see Table.~\ref{tab:canon_symmetry}). For each valid canonical pose, we input it into the simulator to flatten by gravity, which constitutes a predicted goal set. The reward is computed as negative of smallest distance to goals in goal set. We use pairwise l2 distance.

\begin{algorithm*}
\SetAlgoLined
\DontPrintSemicolon 
 \SetKwInOut{Input}{input}
 \SetKwInOut{Output}{output}
 \Input{Depth Image $D$, partial point cloud $P$, mesh reconstruction Model $\phi$, dynamics GNN $G_{dyn}$, number of sampled actions $K$} 
 \Output{pick-and-place action $a = \{x_{pick}$, $x_{place}\}$}
\textbf{Estimate the full mesh of clothes by mesh reconstruction model}: $\tilde{M
}_{init} = \phi(D)$ \\
\textbf{Perform test-time fine-tuning}:  $\tilde{M}_{tuned}^0= finetune(\tilde{M}_{init})=(\tilde{V}^0, \tilde{E}_M) $.\

 \For{$i \gets 1$ \textbf{to} $K$} 
 {

    Sample a pick-and-place action $x_{pick}$, $x_{place}$ \;  
    Compute low-level actions $\Delta x_1, ..., \Delta x_H$\; 
    Get picked point $v_{picked}$ from $x_{pick}$\;
    Pad historic velocities with 0:
    $\mathbf{x}_0 \gets \tilde{V}^0, \dot{\mathbf{x}}_{-m...0} \gets \mathbf{0}$\;
    \For {$t \gets 0$ \textbf{to} $H$}
    {
        Build collision edges $E_C^t$ with $\mathbf{x}_t$\;
        Move  picked point according to gripper movement by : \; 
        $x_{u, t}\gets x_{u, t} + \Delta x_t$, \hspace{0.5pt} $\dot{x}_{u, t} \gets \Delta x_t / \Delta t $\;
        Predict accelerations using $G_{dyn}$:
        $\ddot{\mathbf{x}}_{t} \gets G_{dyn} (\mathbf{x}_t, \dot{\mathbf{x}}_{{t-m}...t}, E_M, E_C^t)$\;
        Update point cloud predicted positions \& velocities: \;
        $\dot{\mathbf{x}}_{t+1} = \dot{\mathbf{x}}_{t} + \ddot{\mathbf{x}}_{t}  \Delta t$, \hspace{0.5pt}
        $\mathbf{x}_{t+1} = \mathbf{x}_{t} + {\dot{\mathbf{x}}}_{t+1}  \Delta t$\;
         Readjust picked point according to gripper movement  by \;
         $x_{u, t}\gets x_{u,t} + \Delta x_t$, \hspace{0.5pt} $\dot{x}_{u, t} \gets \Delta x_t / \Delta t $\;
    }
    Compute reward $r$ based on final mesh nodes position $\mathbf{x}_H$\;

 }
 \Return{pick and place action with maximal reward}\;
 \caption{Planning pipeline of MEDOR}
 \label{algo:planning}
\end{algorithm*}

\section{Implementation Details of Baselines}
\subsection{VisuoSpatial Forsight (VSF)}
We use the official codebase of VSF\footnote{https://github.com/ryanhoque/fabric-vsf}.
Image: we train the model with RGB-D images of size 56 x56 pixels, according to the original paper. To reproduce the performance of the original paper, we collected 7115 trajectories with 15 pick-and-place actions for each category. In total, it amounts to ~100,000 environment steps, which is 5 times the data compared to our methods. 

\textbf{Action sampling} We use a similar action sampling strategy as MEDOR for VSF, that is, bounding box sampling~\ref{sup:medor:plan}. We use a smaller padding size (6 pixels) because of the smaller image size.

\textbf{Reward computation} For flattening, we use color thresholding to compute the coverage of cloth. For canonicalization, we project the predicted and goal RGB-D image into 3D rgb point cloud. The RGB values are scaled to be similar to the coordinate values. Then we run ICP~\cite{zhang1994iterative} for 5 iterations to align the predicted point cloud with the goal point cloud.

\subsection{Visible Connectivity Dynamics (VCD)}
We use the official code of VCD, with modifications so that it works well on our dataset. 

Given point cloud and mesh, the original VCD conduct bipartite matching to map point cloud to mesh nodes. If the corresponding mesh nodes are connected by mesh edges, we also construct mesh edges for the point cloud points. We found that this approach is highly sensitive to density of point cloud and mesh. Imagine the mesh is denser than the point cloud, there might be many mesh vertices between two adjacent points on point cloud. Thus they are not connected although they should. 

To solve this issue, we design a more robust approach for mesh edges construction that is agnostic to the density of mesh. First, for each point on the point cloud, we find the nearest mesh vertex. Then we compute the distance between neighboring points on point cloud by the pairwise geodesic distance of corresponding mesh vertices. A mesh edge is constructed if the geodesic distance is below a threshold.

\textbf{Training} To makes a fair comparison, we train the edge GNN dataset of different categories separately. Each dataset contains 20,000 envrironment steps, which is same as the dataset used for training the mesh reconsctruction model of MEDOR. For dynamics model, similar to MEDOR, we train a single model on Trousers dataset but use it for all categories at test-time.

\textbf{Planning} Same as MEDOR, except that we run ICP~\cite{zhang1994iterative} for 5 iterations to align the predicted point cloud with the goal point cloud before computing the cost function for canonicalization task.


\newpage

\end{document}